\documentclass{article}

\usepackage{microtype}
\usepackage{graphicx}
\usepackage{subfigure}
\usepackage{booktabs} 

\usepackage{hyperref}

\usepackage[accepted]{icml2024}
\makeatletter
\renewcommand{\Notice@String}{}   
\makeatother
\usepackage{amsmath}
\usepackage{amssymb}
\usepackage{mathtools}
\usepackage{amsthm}

\usepackage[capitalize,noabbrev]{cleveref}

\theoremstyle{plain}

\theoremstyle{definition}

\theoremstyle{remark}

\usepackage[textsize=tiny]{todonotes}

\usepackage{dsfont}

\usepackage{multirow}

\usepackage{array}
\newcolumntype{?}[1]{!{\vrule width #1}}

\usepackage{nicefrac}

\usepackage{makecell}

\providecommand{\Description}[2][]{\relax}   
\providecommand{\assuming}{\textbf{Assuming }} 

\DeclareMathOperator*{\argmax}{argmax}

\icmltitlerunning{Learning safe, constrained policies via imitation learning}

\begin{document}

\twocolumn[
\icmltitle{Learning safe, constrained policies via imitation learning: \\
Connection to Probabilistic Inference and a Naive Algorithm}



\icmlsetsymbol{equal}{*}

\begin{icmlauthorlist}
\icmlauthor{George Papadopoulos}{equal,yyy}
\icmlauthor{George A. Vouros}{equal,yyy}
\end{icmlauthorlist}

\icmlaffiliation{yyy}{Department of Digital Systems, University of Piraeus, Piraeus, Greece}

\icmlcorrespondingauthor{George Papadopoulos}{georgepap@unipi.gr}
\icmlcorrespondingauthor{George A. Vouros}{georgev@unipi.gr}

\icmlkeywords{Safe Reinforcement Learning, Constrained Imitation Learning, Dual Gradient Descent.}

\vskip 0.3in
]



\printAffiliationsAndNotice{\icmlEqualContribution} 

\begin{abstract}
This article introduces an  imitation learning method for learning  maximum entropy policies that comply with constraints demonstrated by expert trajectories executing a task. The formulation of the method takes advantage of  results connecting performance to bounds for the KL-divergence between demonstrated and learned policies, and its objective is rigorously justified through a connection to a probabilistic inference framework for reinforcement learning, incorporating the reinforcement learning objective and the  objective to abide by constraints in an entropy maximization setting. The proposed algorithm optimizes the learning objective with dual gradient descent, supporting effective and stable training. Experiments show that the proposed method can learn effective policy models for constraints-abiding behaviour, in settings with multiple constraints of different types, accommodating different modalities of demonstrated behaviour, and with abilities to generalize.
\end{abstract}

\section{Introduction}
\label{sec:Introduction}

In this paper we formulate the problem of learning safe policies with respect to constraints as an imitation learning problem, given expert demonstrations, and we propose a practical algorithm that solves this problem. Specifically, the objective is to learn policies that maximize the discounted accumulated reward of performing a task with respect to the demonstrated constraints-abiding task executions. Costs for violating constraints, or constraints-related rewards, are not revealed explicitly or  approximated, as done for instance in an inverse learning setting, but the learned policy  minimizes the difference between the own occupancy measure and the occupancy measure of the unknown expert policy demonstrated, maximizing its entropy. 
The formulation of the problem is connected to reinforcement learning (RL) as probabilistic inference, providing useful insights and a rigorous justification of the imitation learning objective towards learning safe, constraints-abiding policies, in an entropy maximization setting.

The proposed algorithm implements the dual gradient descent method, where by means of a Lagrangian relaxation method  it maximizes the RL objective (accumulated reward of the learnt policy)  with respect to the demonstrated trajectories adhering to specific constraints. Working in the entropy maximization realm, the SAC algorithm has been chosen due to its ability to tune the entropy coefficient of the learnt policy, while the dual gradient descent method aims to comply with the demonstrated constraints taking into account the policy entropy. Overall, this implements a \textit{double} dual gradient descent method, which aims at maximizing the policy entropy while achieving the reinforcement learning objective, tuning the degree at which the RL objective and policy entropy are maximized  while learning to imitate with respect to the constraints.

The contributions made in this paper are as follows:
\textbf{(a)} The  problem of learning a constraints-abiding policy with respect to demonstrated trajectories is formulated as an imitation learning problem, whose objective function is rigorously justified  through the connection to probabilistic inference \cite{RLasProbInf}.
\textbf{(b)} A practical algorithm of solving the constrained imitation problem is proposed, based on a Lagrangian relaxation  formulation of the imitation problem, with an entropy maximization objective.
\textbf{(c)} The algorithm is evaluated in settings with constraints of increasing complexity, using different modalities of expert behaviour, and is compared to state of the art algorithms.

\section{Preliminaries and Problem specification} \label{sec:preliminaries}
A Markov Decision Process (MDP) is a tuple $(S,A,R,P,\mu)$, where $S$ is the set of states, $A$ is the set of actions available to an agent, $r: S \times A \rightarrow \mathbb{R}$ is the reward function,  $P: S \times A \times S \rightarrow [0,1]$ is the transition probability function $P(s_{t+1}|a_t,s_t)$ to a new state $s_{t+1}$ after the execution of action $a_t$ at state $s_t$, and $\mu: S \rightarrow [0,1]$ is the probability distribution over starting states. 
In such a setting, a learning agent aims to learn a policy $\pi: S \rightarrow P(A)$ from states to probability distributions over actions, so as to maximize the performance measure 
\[J(\pi)=\mathbb{E}_{\tau \sim \pi} [\sum_{t=0}^\infty \gamma^tr(s_t, a_t)]\] 
where $\gamma \in (0,1]$ is the discount factor and $\tau$ denotes any trajectory generated using the policy $\pi$, with $s_0 \sim \mu, a_t \sim \pi(\cdot|s_t), s_{t+1} \sim P(\cdot|s_t, a_t)$.

A constrained Markov decision process (CMDP) is an MDP with constraints that restrict the set of feasible policies for the MDP. Thus, the MDP is augmented with a set of cost functions $C=\{C_1, C_2, ..., C_m\}$, and corresponding cost limits $d_1, d_2, ..., d_m$.  Each $C_i:S \times A  \rightarrow \mathbb{R}$ maps the execution of actions at states to costs.  In such a setting the set of feasible policies with respect to constraints is 
$\Pi_C=\{\pi \in \Pi | \forall i, J_{C_i}(\pi)\leq d_i\}$,
\noindent where  
\begin{equation*}
    J_{C_i}(\pi)=\mathbb{E}_{\tau \sim \pi} [\sum_{t=0}^\infty \gamma^t C_i(s_t, a_t)]
\end{equation*}
 denotes the $i$-th constraint related discounted cost of policy $\pi$ considering the cost function $C_i$. The reinforcement learning (RL) objective is learning the optimal policy
\begin{equation*}
    \pi^*=\argmax_{\pi \in \Pi_C} J(\pi).   
\end{equation*}
The inverse reinforcement learning (IRL) problem aims at fitting a cost function $\tilde{C_i}$, given a set of trajectories generated by an expert policy $\pi_E$, solving the optimization problem 
\[maximize_{\tilde{C_i}}(J_{\pi \in \Pi_{\tilde{C_i}}}(\pi)-J_{\tilde{C_i}}(\pi_E)).\] 
Thus, IRL looks for a cost function that assigns low cost to the expert policy, which must be  feasible,  and high cost to any of the other feasible policies. In doing so, the expert policy can be modeled using a reinforcement learning method 
\begin{equation*}
    \tilde{\pi_E}=\argmax_{\pi \in \Pi_{\tilde{C}}} J(\pi)    
\end{equation*}
where, $\tilde{C}$ is the set of approximated cost functions $\tilde{C_i}$.
 

Fitting these cost functions and evaluating at every approximation the reinforcement learning objective is costly. Therefore, we want to skip this cost-approximation process and, as done for instance in \cite{GAIL}, imitate exploiting the expert demonstrations provided.

Casting the problem in an imitation learning setting we exploit the occupancy measure $\rho_\pi: S \times A \rightarrow \mathbb{R}$, which specifies the distribution of state-action pairs that an agent visits when navigating the environment with policy $\pi$, and is defined (e.g. in  \cite{GAIL}) to be 
\[\rho_\pi(s,a)=\pi(a|s)\sum_{t=0}^\infty \gamma^tP(s_t=s|\pi).\]
Then, the expected cost with respect to a constraint $C_i$ when behaving using $\pi$ is  $\mathbb{E}[C_i(s,a)]=\sum_{s,a}\rho_\pi(s,a)C_i(s,a)$.

Then, as shown in \cite{MDPsBook} the set of valid occupancy measures can be written as a feasible set of affine constraints 
\begin{equation*}
\begin{split}
    D=\{&\rho: \rho \geq 0 \text{ and } \\
    &\sum _a\rho(s,a)=\mu(s)+\gamma\sum_{s',a}P(s|s',a)\rho(s',a), \\
    &\forall s\in S\}
\end{split}
\end{equation*}
The evaluation of these constraints  is  an inefficient process and requires knowledge of the world model. As it is known from (\cite{AppLearnwLP}, Theorem 2) and (\cite{GAIL}, Proposition 3.1), given a $\rho \in D$, then that occupancy measure is associated to a unique policy $\pi_\rho=\rho(s,a)/\sum_{a'}\rho(s,a')$ and this policy is the unique policy  with occupancy measure $\rho$. 

This result, together with the fact that the $\gamma-$discounted causal entropy $\tilde{H}=-\sum_{s,a}\rho(s,a)(log(\rho(s,a)/\sum_a'\rho(s,a'))$ for occupancy measures is strictly concave (\cite{GAIL}, Lemma 3.1), and for all $\pi \in \Pi $ and $\rho \in D$, it holds that $\mathcal{H}(\pi)=\tilde{\mathcal{H}}(\rho_\pi)$ and  $\mathcal{H}(\pi_\rho)=\tilde{\mathcal{H}}(\rho)$,   allow us to switch between policies and occupancy measures when considering functions involving causal entropy and expected costs, as it follows: \[\mathbb{E}_\pi[C_i(s,a)]-\mathcal{H}(\pi)=\sum_{s,a}\rho(s,a)C_i(s,a)-\tilde{\mathcal{H}}(\rho).\] This holds for any cost function, occupancy measure $\rho \in D$ and corresponding policy $\pi=\pi_\rho$, as well as, for any policy $\pi$ and the corresponding occupancy measure $\rho=\rho_\pi$. 

Therefore, to solve the imitation learning problem given samples from a policy $\pi_E$, we need to find a maximal entropy policy $\pi$ that minimizes the difference between its own occupancy measure and the occupancy measure of  $\pi_E$. 
Intuitively, in a constrained setting, this implies that high cost state-actions will be visited as often as it is done by the expert policy, and thus they will incur the implied cost according to the visitation probabilities, while the high entropy ensures that the agent will explore to model a policy that generalizes sufficiently well. 

Given this objective, the  imitation learning problem  of  maximizing $J(\pi)$ and minimizing the divergence between $\rho_{\pi_{E}(s,a)}$ and $\rho_\pi(s,a)$, with respect to the demonstrations provided from the expert policy $\pi_E$, can be formulated  as follows:
\[J(\pi)=\mathbb{E}_{\tau \sim \pi} [\sum_{t=0}^\infty \gamma^tr(s_t, a_t)]\] 
\begin{equation} 
\label{eq:max_min}
\begin{split}
    &\max_{\pi} J(\pi)+\mathcal{H}(\pi) \\
    & \mathrm{s.t.} \ Dist_\rho(\pi_E, \pi) \leq \delta \textit{, given } \mathcal{D}
\end{split}
\end{equation}
where, $\mathcal{D}$ denotes the set of expert samples provided, $Dist_\rho(\pi_E, \pi)$ is a distance measure applied on occupancy measures and $\delta \geq 0$ limits how closely the two policies should be, in terms of occupancy measures. 
Objective (1) can be used for problems with soft (given a $\delta \geq$ 0) and hard constraints (given $\delta = 0$).

This objective ensures  (a) that the learnt policy will be nearly (given that small divergences from the expert policy may incur high costs) as feasible as the expert policy, and  (b) the expected return will be at least as good as that of the expert policy.
The latter is ensured by the policy performance bounds, in terms of costs or rewards, proved in \cite{CPO} between two stochastic policies: The difference in expected return or expected cost between two stochastic policies, $\pi$ and $\pi_E$, is bounded by their KL divergence (denoted $D_{KL}$) in all visited states, given that states are sampled according to the discounted future state distribution $d_{\pi_E}(s)=\frac{1}{1-\gamma}\sum_{t=0}^\infty \gamma^tP(s_t=s|\pi_E)$ by following $\pi_E$. 

Therefore, the objective (1) can be stated in an empirical way in terms of KL divergence between policies, as follows:
\begin{equation} 
\label{eq:max_min}
\begin{split}
   max_{\pi} [& ( J(\pi)+\mathcal{H}(\pi) ) - \\
   & ( \mathbb{E}_{s \sim d^{\pi_E}} [ \mathbb{E}_{a \sim \pi_{E}}[D_{KL}(\pi_E || \pi)(s)] ] - \delta )]  \approx \\
   max_{\pi} [& ( J(\pi)+(1-\beta)\mathcal{H}(\pi) ) - \\
   & ( \mathbb{E}_{s \sim d^{\pi_E}} [ \mathbb{E}_{a \sim \pi_{E}}[D_{KL}(\pi_E || \pi)(s)] - \beta \mathcal{H}(\pi)] - \delta )]
\end{split} 
\end{equation}
where, the entropy term is split by a factor $\beta \in [0,1]$ to the RL objective and the  objective to abide by the constraints. The factor $\beta$  balances the entropy term between optimizing the term $\max_{\pi} J(\pi)$ and imitating the expert policy, i.e. complying with constraints. Intuitively, $\pi$ must minimize its distance from $\pi_E$, taking also into account the entropy term. This has  important consequences in the algorithmic part and is  further discussed below.
As a remark, given that the agent does not have access to $\pi_{E}$, we cannot calculate $D_{KL}(\pi || \pi_E)$, but  we can calculate $D_{KL}(\pi_E || \pi)$ to evaluate the policy $\pi$ using the provided demonstrations from $\mathcal{D}$. This also has the effect of learning to behave according to different behavior modalities demonstrated by $\pi_E$.

As known, $D_{KL}(P(x) || Q(x)) = \mathbb{E}_{x \sim P(x)}[-logQ(x)] - \mathcal{H}(P(x))$ where $P(x)$ and $Q(x)$ are arbitrary distributions, $x$ is a random variable, and $\mathcal{H}(P(x)) = E_{x \sim P(x)}[-logP(x)]$ is the entropy of $P(x)$. Considering that $P(x)$ is the true distribution, then minimizing $D_{KL}(P(x) || Q(x))(s)$, will force $Q(x)$ to be as close as possible to $P(x)$. In our case, where we aim to impel agent policy $\pi$ close to the constraint-compliant policy $\pi_{E}$, we need to minimize:
\begin{equation} \label{eq:d_kl}
\begin{split}
    &\mathbb{E}_{s \sim d^{\pi_E}, a \sim \pi_E}[D_{KL}(\pi_E || \pi)(s)] = \\
    &\mathbb{E}_{s \sim d^{\pi_E}} [E_{a \sim \pi_E}[-\log(\pi(a | s))] - \mathcal{H}(\pi_E(\cdot | s))]= \\
     &\mathbb{E}_{s \sim d^{\pi_E}} [E_{a \sim \pi_E}[-\log(\pi(a | s))]
\end{split}
\end{equation}
The last equality holds, considering that the demonstrated policy can be assumed deterministic, especially when this concerns few demonstrations in a very large state-action space. The  objective (\ref{eq:max_min}) is also justified by connecting the constraints-abiding behaviour imitation problem to the reinforcement learning as probabilistic inference framework proposed in \cite{RLasProbInf}, by incorporating  the constraints satisfaction probabilities into the formulation. This is shown in the Appendix \ref{sec:ImitationAsProbInf}.

\section{Safe and constrained imitation learning}

This section specifies the dual gradient descent problem formulation by means of the Lagrangian relaxation, and it specifies the practical algorithm to solve the constraints-abiding  imitation problem.

\subsection{Problem formulation} 
\label{sec:prob_form}
We consider a set $\mathcal{D}=\{(s,a)_1, (s,a)_2, ..., (s,a)_n\}$ of demonstrations collected following policy $\pi_{E}$, i.e., 
states and  actions at any timestep $t$. In addition, we consider a set of constraints $\mathcal{C}=\{c_1, c_2, ..., c_m\}$ corresponding to cost functions $\{C_1, C_2, ..., C_m\}$. The constraints must be satisfied when performing the task, thus the  demonstrator acts with respect to $\mathcal{C}$ without explicitly specifying them. Furthermore, $\pi_{E}$ is such that the task is executed in a near-optimal way (i.e. optimal with respect to constraints) in terms of the RL objective. 

Our ultimate goal is to train an agent to exhibit a constraints-abiding optimal behavior, i.e., to perform the demonstrated task  satisfying the  demonstrated constraints, while acting as optimally as possible. During training, the agent has only access to the collected demonstrations and it receives a reward after executing an action $a$ in the current state $s$. The reward is decoupled from constraint adherence, and it is possible for the constraints to be at odds with the provided reward.




Following previous works \cite{Altman1999ConstrainedMD, CPO, pmlr-v119-stooke20a}, a feasible direction to solve the above \textit{max-min} problem is to transform the objective into a  Langrange dual function using Lagrangian relaxation. 
As in \cite{Boyd2004}, and with regard to Eq. (\ref{eq:max_min}) we  define the dual function:
\begin{equation}
\begin{split}
\label{eq:lag_relax}
    \ \mathcal{L}(\pi, \lambda) = &-[J(\pi)+ (1-\beta) \mathcal{H}(\pi)] \\
    &+ \lambda [D_{KL}(\pi_E || \pi)- \beta \mathcal{H}(\pi) - \delta]  
\end{split}
\end{equation}
where $\lambda > 0$ denotes the Lagrangian multiplier. 

Then, we optimise the Lagrangian dual function :
\begin{equation} \label{eq:dual_func}
\begin{split}
    & g(\lambda) = \mathcal{L}(\pi^*, \lambda) \\
    & \mathrm{s.t.} \ \pi^* = \mathrm{arg}\min_{\pi} \mathcal{L}(\pi, \lambda) 
\end{split}    
\end{equation}
Considering that $\pi$ can be expressed as a parameterized policy with parameters $\theta$, denoted by $\pi_{\theta}$, we can use a truncated \footnote{Similarly to \cite{haarnoja2019soft}, we can approximate dual gradient descent by iteratively updating each objective using a single gradient step instead of first optimizing to convergence $\mathcal{L}(\pi_{\theta}, \lambda)$ with respect to $\pi_{\theta}$ and then $g(\lambda)$ with respect to $\lambda$.} version of the dual gradient descent method to optimize both $\pi_{\theta}$ and $\lambda$. In doing so, we iteratively perform two update steps until convergence: First, minimize $\mathcal{L}(\pi_{\theta}, \lambda)$ with respect to $\pi_{\theta}$ keeping $\lambda$ fixed. This yields the optimal $\pi_{\theta^*}$. Then, maximize $g(\lambda)$ with respect to $\lambda$ keeping $\pi_{\theta^*}$ fixed. This process results at a saddle point between $\max [J(\pi)+ (1-\beta) \mathcal{H}(\pi)]$ (the RL objective term) and $\min \lambda [D_{KL}(\pi_E || \pi) - \beta \mathcal{H}(\pi) - \delta]$ (the constraints term). The interpretation is that we find a policy $\pi_{\theta}$ that softly satisfies the constraint set $\mathcal{C}$ while maximizing as much as possible $J(\pi_{\theta})$. It must be noted that, as it is well known \cite{liu2021policy}, the Lagrangian relaxation method is sensitive to the initialization of the Lagrangian multipliers.


Finally, it must be noted that incorporating the entropy term in both terms in Eq. (\ref{eq:lag_relax})
is very important. Specifically, with regards to the maximum entropy principle \cite{Ziebart2010Modeling}, and based on the observations of \cite{pmlr-v119-stooke20a}, 
there is an inherent drawback of Lagrangian relaxation which makes $g(\lambda)$ and $\lambda$ to oscillate with fixed phase, making the learning process unstable. In the case of the maximum entropy policy, this issue is intensified since, the maximization of the RL objective softly forces the action distribution to balance between the maximum rewards and the maximum entropy, while the minimization of the constraints term of Eq. (\ref{eq:lag_relax}), i.e., $\lambda [D(\pi_E, \pi) - \beta \mathcal{H}(\pi) - \delta]\textit{, with }\beta =0$, combined with the maximization of $g(\lambda)$ with respect to $\lambda$,  will strictly force the policy $\pi$ towards choosing actions indicated by the demonstrated expert policy $\pi_E$. 
%


\subsection{The SCOPIL algorithm}
This section presents a practical algorithm for implementing the  dual gradient descent method described in the previous section. 
Our algorithm is composed of 3 main components: (a) an RL algorithm, (b) a constraints' adherence component, and (c) a component for optimizing (minimizing or maximizing according to the constraint term value) the Lagrange multiplier.

Starting with (a), we chose SAC \cite{pmlr-v80-haarnoja18b} since it is a maximum entropy state-of-the-art method for learning stochastic policies, with the ability to self-tune the policy entropy coefficient, which has interesting implications to satisfying the overall objective (discussed below). In this work we use the discrete version of SAC\footnote{Our implementation of discrete SAC is based on the implementation of \cite{10.1145/3453892.3454004}. URL: \url{https://github.com/ligerfotis/maze_RL_v2}} presented in \cite{christodoulou2019soft}. In brief, given a parameterized policy $\pi_{\theta}$ and a replay buffer $\mathcal{B}$ where state-action $(s,a)$ pairs from policy rollouts are stored, the overall objective  to minimize is the following:
\begin{align}
\label{eq:SAC_pol_obj}
    J_{\pi}(\theta) &= \mathbb{E}_{s \sim \mathcal{B}} \big[ - \pi_{\theta}(s)^T [\alpha \log(\pi_{\theta}(s)) \notag \\
    &\phantom{= \mathbb{E}_{s \sim \mathcal{B}} \big[ - \pi_{\theta}(s)^T [}\; - \min(Q_{\phi_1}(s), Q_{\phi_2}(s))] \big] \notag \\
    &= \mathbb{E}_{s \sim \mathcal{B}} \big[ - \alpha \pi_{\theta}(s)^T \log(\pi_{\theta}(s)) \notag \\
    &\phantom{= \mathbb{E}_{s \sim \mathcal{B}} \big[ }\; + \pi_{\theta}(s)^T \min(Q_{\phi_1}(s), Q_{\phi_2}(s)) \big] \notag \\
    &= \mathbb{E}_{s \sim \mathcal{B}} \big[ \alpha \mathcal{H}(\pi_{\theta}(\cdot | s)) \notag \\ 
    &\phantom{= \mathbb{E}_{s \sim \mathcal{B}} \big[}\; + \pi_{\theta}(s)^T \min(Q_{\phi_1}(s), Q_{\phi_2}(s)) \big]
\end{align}
$\mathcal{H}(\pi_{\theta}(\cdot | s))$ is  an approximation of the policy entropy based on the samples stored in $\mathcal{B}$. The term $\alpha$ is the learnable entropy coefficient. 
The objective to maximize for updating $\alpha$ 
is the following:
\begin{equation} \label{eq:entr_coef_obj}
\begin{split}
    J(\alpha) &= \mathbb{E}_{s \sim \mathcal{B}} \big[ \alpha [\Bar{\mathcal{H}} - \pi_{\theta}(s)^T \log(\pi_{\theta})] \big] \\
    &= \mathbb{E}_{s \sim \mathcal{B}} \big[ \alpha [\Bar{\mathcal{H}} - \mathcal{H}(\pi_{\theta}(\cdot | s))] \big]
\end{split} 
\end{equation}
where $\Bar{\mathcal{H}}$ is the target entropy, set equal to $0.4 \log |A|$, according to \cite{10.1145/3453892.3454004}, given the set $A$ of agent's discrete actions. In addition, $Q_{\phi_1}(s)$ and $Q_{\phi_2}(s)$ in Eq. (\ref{eq:SAC_pol_obj}) are approximated by neural networks parameterized by $\phi_1$ and $\phi_2$, respectively. 
Each one of these is updated using the following objective (to minimize):
\begin{equation} 
\label{eq:J_Q} 
\begin{split}
    &J_Q(\phi_x) = \\
    &\mathbb{E}_{(s,a,s') \sim \mathcal{B}} \big[ \frac{1}{2} \big( Q_{\theta_x}(a,s) - (R(s,a,s') + \gamma V_{\bar{\phi}_{1,2}}(s')) \big)^2 \big], \\
    &\forall x \in \{1,2\}  
\end{split}
\end{equation}
where $V_{\bar{\phi}_{1,2}}(s')$ is a practical sample-based approximation of $ \mathbb{E}_{s' \sim P(s,a)}[V_{\bar{\phi}_{1,2}}(s')]$ calculated using two target networks, $Q_{\psi_1}$ and $Q_{\psi_1}$. Specifically:
\begin{equation}
\begin{split}
    &V_{\bar{\phi}_{1,2}}(s) = \\
    &\mathbb{E}_{a \sim \pi_{\theta}(s)} [\min(Q_{\psi_1}(s), Q_{\psi_2}(s)) - \alpha \log(\pi_{\theta}(a \mid s))] = \\
    &\pi_{\theta}(s)^T [\min(Q_{\psi_1}(s), Q_{\psi_2}(s)) - \alpha \log(\pi_{\theta}(s))] = \\
    &\pi_{\theta}(s)^T \min(Q_{\psi_1}(s), Q_{\psi_2}(s)) - \alpha \mathcal{H}(\pi_{\theta}(s))
\end{split}
\end{equation}
On their turn, $Q_{\psi_1}$ and $Q_{\psi_1}$ are softly updated using the exponential average of $Q_{\phi_1}$ and $Q_{\phi_1}$ parameters:
\begin{equation} \label{eq:q_soft_update}
\begin{split}
    \psi_x' = \mathcal{\epsilon} \phi_x + (1-\mathcal{\epsilon}) \psi_x, \forall x \in \{1,2\}  
\end{split}
\end{equation}
where $\epsilon$ is a hyperparameter. 

Incorporating the entropy term $\alpha$ in Eq. \ref{eq:lag_relax} , and considering the parameterized policy $\pi_\theta$, this becomes

\begin{equation}
\begin{split}
\label{eq:SAC_lag_relax}
    \ \mathcal{L}(\pi_\theta, \lambda) = &-[J_\pi(\theta)+ (1-\beta) \alpha \mathcal{H}(\pi_\theta)] \\
    &+ \lambda [D_{KL}(\pi_E || \pi_\theta)- \beta \alpha \mathcal{H}(\pi_\theta) - \delta]  
\end{split}
\end{equation}

Incorporating the learnable coefficient $\alpha$ in both terms in Eq. \ref{eq:SAC_lag_relax}, encourages the agent to explore for maximizing the RL objective, considering also the effect of the entropy term (weighted by $\alpha$) to comply with the constraints. To avoid non-learnable hyperparameters and considering that (a) the entropy term contributes equally to both terms of the objective, thus $\beta=0.5$, and (b)  $\alpha$ can be tuned to any value that can be scaled by $1/\beta=2$ in both terms when $\beta=0.5$, Eq. (\ref{eq:SAC_lag_relax}) becomes
\begin{equation} 
\label{eq:lagr_algo}
\begin{split}
    &\mathcal{L}(\pi_{\theta}, \lambda) = \\ 
    &- \big( \mathbb{E}_{s \sim \mathcal{B}} \big[ \alpha \mathcal{H}(\pi_{\theta}(\cdot | s)) + \pi_{\theta}(s)^T \min(Q_{\phi_1}(s), Q_{\phi_2}(s)) \big] \big) \\
    &+ \lambda \Big( \mathbb{E}_{s \sim \mathcal{D}} \big[ \mathbb{E}_{a \sim \mathcal{D}} [-\log\big(\pi_\theta(a | s) \big) ] - \alpha \mathcal{H}\big( \pi_{\theta} (\cdot | s) \big) \big] - \delta \Big)
\end{split}
\end{equation}

where $J_\pi(\theta)$  has been substituted from Eq. (\ref{eq:SAC_pol_obj}).

As it can be noticed, with regard to the constraints adherence component, we use a sampled-based approximation of Eq. (\ref{eq:max_min}), where $(s,a)$ pairs are sampled from the set $\mathcal{D}$. 

Regarding the optimization of the Langrangian multiplier $\lambda$ we 
maximize $g(\lambda)$:
\begin{equation} 
\label{eq:g_algo}
\begin{split}
    &g(\lambda) = \mathcal{L}(\pi_{\theta^*}, \lambda) = \\
    &\lambda \Big( \mathbb{E}_{s \sim \mathcal{D}} \big[ \mathbb{E}_{a \sim \mathcal{D}} [-\log\big( \pi_{\theta^*}(a | s) \big) ] - \alpha \mathcal{H}\big( \pi_{\theta^*} (\cdot | s) \big) \big] - \delta \Big)   
\end{split}
\end{equation}

where, the  $-J_{\pi}(\theta)$ term is omitted since, it does not contribute to the gradients of $g(\lambda)$ with respect to $\lambda$. 

 We summarize the proposed method in Algorithm (\ref{algo:scopil}).

\begin{algorithm}[hbt!]
\caption{SCOPIL} \label{algo:scopil}
\begin{algorithmic}
\STATE Collect expert demonstrations set: $\mathcal{D}$
\STATE Initialize neural networks parameters: $\theta$, $\phi_1$, $\phi_2$, $\psi_1$, $\phi_2$
\STATE Initialize Lagrangian multipliers: $\lambda$, $\alpha$ \\
\STATE Initialize an empty replay buffer: $\mathcal{B} \gets \varnothing$
\STATE /* $\eta$ and $\kappa$ are the learning rates*/
\FOR {each episode}
    
    \FOR {each environment step}
     \STATE $a \sim \pi_{\theta}(a|s)$
     \STATE $s' \sim P(s'|s,a)$
     \STATE $\mathcal{B} \gets \mathcal{B} \bigcup \{s,a,R(s,a,s'),s'\}$
     \ENDFOR

    \FOR {each gradient step}
    \STATE $\phi_x \gets \phi_x - \mathcal{\eta} \hat{\nabla}_{\phi_x} J_Q(\phi_x)$, $\forall x \in \{1,2\}$ \null\hfill Eq.(\ref{eq:J_Q})
    \STATE $\theta \gets \theta - \mathcal{\eta} \hat{\nabla}_{\theta}\mathcal{L}(\pi_{\theta}, \lambda)$ \null\hfill Eq. (\ref{eq:lagr_algo})
    \STATE $\alpha \gets \alpha - \mathcal{\kappa} \hat{\nabla}_{\alpha} \left( -J(\alpha) \right)$ \null\hfill Eq. ({\ref{eq:entr_coef_obj}})
    \STATE $\lambda \gets \lambda - \mathcal{\eta} \hat{\nabla}_{\lambda} \left( -g(\lambda) \right)$ \null\hfill Eq. (\ref{eq:g_algo})
    \STATE $\psi_x \gets \epsilon \phi_x + (1-\epsilon)\psi_x$, $\forall x \in \{1,2\}$ \null\hfill Eq. (\ref{eq:q_soft_update}) 
    \ENDFOR

\ENDFOR
\end{algorithmic}
\end{algorithm}

\section{Experiments}

\subsection{Experimental settings}

\begin{figure}[h]
  \centering
  \includegraphics[width=0.8\linewidth]{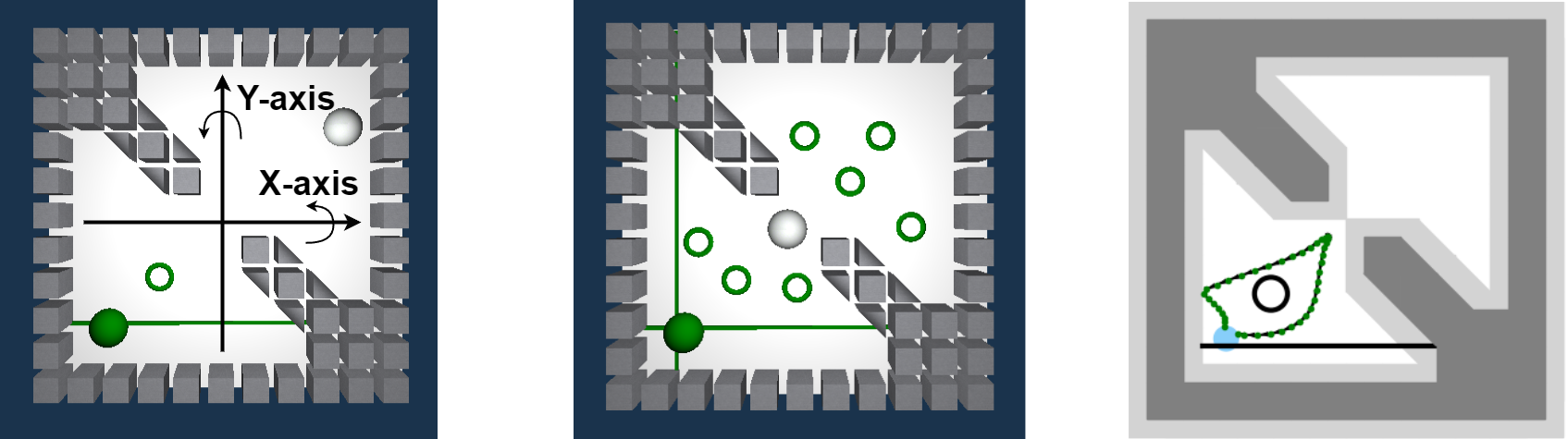}
  \vspace*{-4mm}
  \caption{The Marble Maze environment with constraints. (Left): \textit{Simple}, (Center) \textit{Multiple-constraints}, (Right) \textit{Two-modes}.}
  \label{fig:game_board_multiconstraint_multimodes}
  \Description{}
\end{figure}

To evaluate our method, we use the Marble Maze environment provided by \cite{10.1145/3453892.3454004}. An illustration of the environment is presented in Figure \ref{fig:game_board_multiconstraint_multimodes}(left). The goal is to direct the white ball toward the solid-green hole as fast as possible, by rotating the board in the Y and in the X axes. The environment is augmented with constraints which are depicted with green straight lines and green circles. Specifically, we designed three different settings, \textit{Simple}, \textit{Multiple-constraints}, and \textit{Two-modes}, presented in Figure \ref{fig:game_board_multiconstraint_multimodes}(left, middle, right), respectively. In the \textit{Simple} setting the agent should not drop the ball bellow the green straight line and should not cross the green circle. When one of these cases occurs we consider it as a constraint violation. In the \textit{Multiple-constraints} setting, we introduce seven cyclic constraints, a horizontal and a vertical straight line, setting the constraints. We consider that the vertical constraint is violated when the ball is on the left side of the line. The \textit{Two-modes} setting is identical to \textit{Simple}, except from the provided demonstrations. Here, from each starting point there are two trajectories, one is directed left to the cyclic constraint and one directed right. In this way, we aim to show the ability of SCOPIL to address the well-known mode collapse problem \cite{10.5555/3295222.3295284}.

In each game, the agent should reach the target in 200 timesteps, otherwise the game ends. A state $s$ of the environment is specified by the following vector:
\begin{equation*}
    s = [b_x, b_y, v_x, v_y, r_x, r_y, rv_x, rv_y]
\end{equation*}

\noindent where, $(b_x, b_y)$ is the ball position, $(v_x, v_y)$ is the ball velocity, $r_x$ and $r_y$ are the rotating angles of the board in X and Y axes respectively, while $(rv_x, rv_y)$ are the corresponding rotating velocities of the board. State vectors are normalized in the range $[-1,1]$ using min-max normalization, with $max$ and $min$ values set  empirically.

%
%
%
%
After observing the state, the agent decides to rotate the board by choosing one of the 9 available (X,Y)-joint actions: 
$(No Move, No Move)$, $(No Move, Turn left)$, $(No Move, Turn Right)$, $(Up, No Move)$, $(Up, Turn Left)$, $(Up, Turn Right)$, $(Down, No Move)$, $(Down, Turn Left)$, $(Down, Turn Right)$. The selected action is applied for a specific time period which corresponds to a timestep. The timestep duration is measured in environment updates rather than actual time (e.g. in nanoseconds). Specifically, we update the environment 5 times after applying an action.

At each timestep $t$ the agent receives a goal-related reward, which is defined as follows:
\[
    R(s,a,s') = \left\{
     \begin{array}{@{}l@{\thinspace}l}
       +10 &: \text{if goal reached} \\
       -5 &: \text{if timeout} \\
       \nicefrac{-d(b,h)}{max_{d}} &: \text{otherwise} \\

     \end{array}
   \right.
\]

\noindent where, $d(b,h)$ denotes the Euclidean distance between the ball and the hole, i.e., $\sqrt{(b_x - h_x)^2 + (b_y - h_y)^2}$, and $max_{d}$ is the maximum possible distance. This reward function drives the agent to learn to move the ball and reach the goal state as soon as possible. The reward at each timestep is normalized in the range $[-1,0]$ using min-max normalization
%
%
with $max_R=10$ and $min_R=-5$.

\begin{figure}[h]
  \centering
  \includegraphics[width=0.8\linewidth]{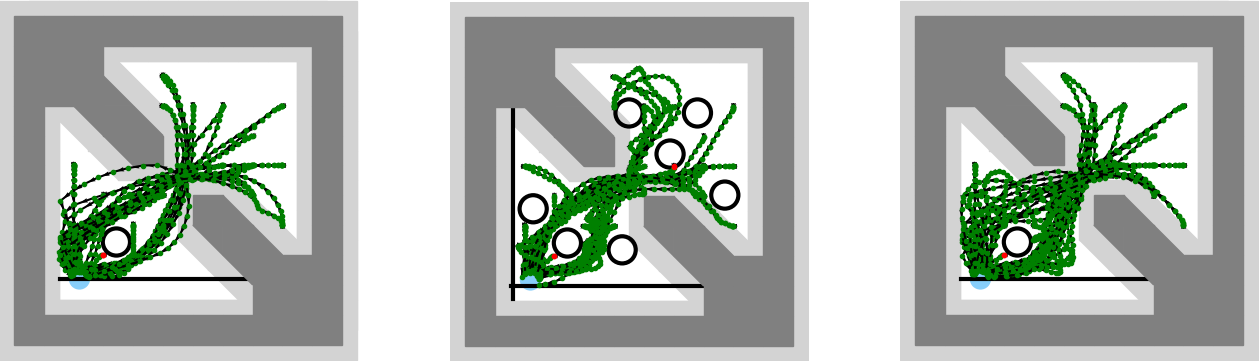}
  \vspace*{-4mm}
  \caption{Collected sets of demonstrations. (Left): \textit{Simple-dem}, (Center) \textit{Multiple-constraints-dem}, (Right) \textit{Two-modes-dem}.}
  \label{fig:demonstrations}
  \Description{}
\end{figure}

For the ease to collect expert demonstrations for each of the three settings, we introduced a ``freeze" function in the game.  The number of games per setting is set to 40, with a random seed regarding the starting ball position. Table \ref{table:demonstrations} summarizes the demonstrations per setting (\textit{Simple-dem}, \textit{Multiple-constraints-dem}, and \textit{Two-modes-dem}), specifying the number of state-action pairs included in each set, the average reward per game, as well as the average length of trajectories and trajectory timesteps (Steps).
 Figure {\ref{fig:demonstrations}} visualizes  the demonstrations. The green coloured points indicate states with no constraint violation,  while red ones are  states  violating a constraint. Note that in \textit{Simple-dem} and \textit{Two-modes-dem} there is one violation beneath the cyclic constraint at the starting state. In \textit{Multiple-constraints-dem} setting, there are 2 such violations.  

\begin{table*}
\begin{center}
\caption{Demonstrations sets specifications}
\footnotesize
  \begin{tabular}{ l | c | c | c | c }

    \Xhline{0.15em}
     \textbf{Demonstrations set} & \textbf{\# state-action pairs} & \textbf{Reward}  & \textbf{Length} & \textbf{Steps} \\
     \Xhline{0.15em}
     
    \textit{Simple-dem} & 850 & -4.50$\pm$9.57 & 256.04$\pm$134.01 & 21.25$\pm$8.62\\ \hline
    \textit{Multiple-constraints-dem} & 1204 & -20.85$\pm$12.02 & 265.10$\pm$138.28 & 30.10$\pm$16.33 \\ \hline
    \textit{Two-modes-dem} & 1067 & -18.30$\pm$8.51 & 280.49$\pm$129.48 & 26.67$\pm$11.69 \\
    \Xhline{0.15em}
  
  \end{tabular}
  \label{table:demonstrations}
\end{center}
\end{table*}

SCOPIL is compared to the state of the art inverse constrained RL (ICRL) method, as well as to SAC in the $Simple$ setting. We do not provide comparative results in the other settings because the superiority of SCOPIL over ICRL is apparent even in the most simple case. Our research questions are as follows:

\textbf{Q1:} \textit{Can SCOPIL yield agents capable of learning and respecting the demonstrated constraints without explicitly modeling them?}

\textbf{Q2:} \textit{Are the agents trained by SCOPIL able to address the mode collapse problem without specifying any information about the modes during training and execution?}

\textbf{Q3:} \textit{How the entropy term in the constraints term of the objective contributes to the efficacy of the method?}

Results are generated by 10 independent, randomly initialized experiments. Each experiment comprises 40 games, starting from initial states of the demonstrated trajectories. Results report on (a) the average total reward accumulated across a trajectory, (b) the average  number of horizontal (H), vertical(V) and  cyclic (C) constraints violations, and on the average number of constraint violations for all types of constraints (e.g. H+V or H+C+V), as well as (c) the average frequency (F(H),  F(V), F(C)) of corresponding constraints violations, and the average frequency of all types of constraint violations (e.g. F(H+V) or F(H+C+V)). The average length of generated trajectories (total spatial distance of the trajectory) and the average number of timesteps (steps) per trajectory are reported. Regarding the  reward, the number of violations, length and steps, we provide their macro average  for each of the 40 games and for the 10 experiments. For the frequency of violations, we divide the total number of violations in a game by the total number of steps, and then we compute the macro average results for all games and experiments. 
Appendix \ref{sec:CurvesandTraj} provides the learning curves of SCOPIL and compared methods, and shows generated trajectories in all settings, indicating states (in red) where constraints are violated. 

SCOPIL is trained for a specific number of steps ($10^6$), which are shown to be sufficient to learn to adhere to the constraints demonstrated. SAC is also trained for $10^6$ steps, while ICRL for $8*10^6$ steps. The number of episodes vary depending on the number of steps per trajectory. 



\subsection{Experimental results}
\label{sec:results}
\textbf{Learning Simple Constraints.} As it is shown in Table \ref{table:simple_setting_results}  SCOPIL achieves exceptional performance, with only 0.045 and 0.042 average number of horizontal and cyclic violations per game, respectively. In total it reports on average 0.087 constraint violations, which is an order of magnitude less than that reported by ICRL (1.54) and SAC (1.99). On the other hand, ICRL exhibits inferior performance yielding 0.52 and 1.02 average number of violations for the horizontal and cyclic constraint, respectively. Furthermore, the generated trajectories illustrated in Figure \ref{fig:results_trajectories_scopil_icrl} in Appendix \ref{sec:CurvesandTraj} indicate that the trained SCOPIL agents not only have learnt to abide by the constraints but also have managed to address mode collapse by maneuvering around the cyclic constraint. The corresponding visualizations for ICRL show that it fails to generate trajectories that respect the cyclic constraint  due to mode collapse. This happens due to the 
binary classifier assessment provided to the policy, which interpolates between the two expert modes. 
SAC does not take into consideration the constraints. Interestingly, SAC manages to satisfy the cyclic constraint, by simply avoiding to collapse the demonstrated modes. The reward achieved by SAC is higher than SCOPIL and ICRL, which is natural given that it does not concern about adhering to constraints. This indicates how SCOPIL manages to compensate the maximization of the reward (RL optimization) while  minimizing the constraint violations by adhering to the demonstrated trajectories. 
The ablation study reported in Appendix \ref{sec:Ablation} (Table \ref{table:simple_setting_results_ablation_study}) reports on SCOPIL with a non-effective dual gradient descent (denoted S-DGD) and with a non-effective dual gradient descent without the entropy in the constraints term (denoted S-DGD-EiC). S-DGD maximizes the objective of SAC with respect to constraints but with a constant Lagrangian multiplier $\lambda=1.05$. The same happens for S-DGD-EiC but without the entropy in the constraints term. Results show that both variations are not as much effective  in satisfying the constraints as SCOPIL, however, they report very good results for the cyclic constraints, due to the reasons explained  with regard to SAC. This is true for S-DGD-EiC for another reason, of close relevance to Q3: It ``forces" close match to the occupancy measure of the demonstrated policy, reducing the ability of the method to generalize beyond samples seen.
This is elaborated further in Appendix \ref{sec:Ablation}.

Figure \ref{fig:main_results} in Appendix \ref{sec:CurvesandTraj}, shows the learning curves of all methods: SCOPIL manages to reduce the frequency of violations and the number of violations significantly faster than the other methods, while the rate of increasing the reward, is similar to that of the other methods.

It must also be noted that the average length and average number of steps of the SCOPIL-generated trajectories, reported in Table \ref{table:simple_setting_results}, are closer to the corresponding measures of the demonstrated trajectories, compared to those reported by the other methods: ICRL trajectories have on average smaller  length, which shows the tendency to avoid maneuvering around the circle, while SAC generates longer trajectories on average due to the maximization of entropy without adhering to the demonstrated trajectories.



\begin{table}
\begin{center}
\caption{Results for SCOPIL, ICRL, and SAC in \textit{Simple} setting.}
\scriptsize
  \begin{tabular}{ l | c | c | c }

    \Xhline{0.15em}
    \textbf{Measure} &  \textbf{SCOPIL} & \textbf{ICRL} & \textbf{SAC} \\
    \Xhline{0.15em}

   {Rwd} & -13.46$\pm$0.29 & -11.88$\pm$0.10 & -11.31$\pm$0.28  \\ \cline{2-4}
    \Xhline{0.15em}
    
    {H } & 0.045$\pm$0.95 & 0.52$\pm$0.60 & 1.11$\pm$0.20 \\ \cline{2-4}
    \Xhline{0.15em}
    
   {F(H) } & 0.002$\pm$0.005 & 0.024$\pm$0.026 & 0.05$\pm$0.01 \\ \cline{2-4}
    \Xhline{0.15em}
    
   {C }  & 0.042$\pm$0.04 & 1.02$\pm$0.23 & 0.87$\pm$0.15 \\ \cline{2-4}
    \Xhline{0.15em}
    
    {F(C) } & 0.004$\pm$0.001 & 0.08$\pm$0.016 & 0.08$\pm$0.01 \\ \cline{2-4}
    \Xhline{0.15em}

    {H+C } & 0.087$\pm$0.126 & 1.54$\pm$0.53 & 1.99$\pm$0,19 \\ \cline{2-4}
    \Xhline{0.15em}
    
    {F(H+C) } & 0.006$\pm$0.006 & 0.10$\pm$0.027 &  0.13$\pm$0.139\\ \cline{2-4}
     \Xhline{0.15em}
     
    {Length} &  255.32$\pm$1.57 & 244.47$\pm$12.73 & 260.05$\pm$4.99 \\ \cline{2-4}
    \Xhline{0.15em}

   {Steps} & 19.83$\pm$0.41 & 17.53$\pm$1.38 & 16.79$\pm$0.40 \\ \cline{2-4}    
    \Xhline{0.15em}
	
  \end{tabular}
  \label{table:simple_setting_results}
\end{center}
\end{table}

\textbf{Learning Complex Constraints.} Scaling up the complexity of the setting, SCOPIL still manages to maintain the average number of violations in low levels using the same number of demonstrated trajectories when tested in train seed starting points. As shown in Table \ref{table:multiconstraints_setting_results}, SCOPIL manages to score an average reward (-21.84) very close to that of the demonstrated trajectories (-20.85) shown in Table \ref{table:demonstrations},  while the average number of horizontal, vertical, and cyclic constraints violations are 0.81, 0.55, and 0.35, respectively (0.57 for all three types of constraints). This behavior shows the ability of SCOPIL to handle various types of constraints that present trade-offs, in the sense that the satisfaction of one of these can make the satisfaction of the others harder. 
Training curves for SCOPIL on the $Multiple-constraints$ setting are shown in Figure \ref{fig:scopil_results_multiconstraints} in Appendix \ref{sec:CurvesandTraj}, together with trajectories generated by SCOPIL in Figure \ref{fig:multiconstraints_scopil_all_trajectories_scopil_only_4}: These show the effectiveness of the method to reduce violations while increasing the reward.


\begin{table}
\begin{center}
\caption{Results for SCOPIL in \textit{Multiple-constraints} setting.}
\scriptsize
  \begin{tabular}{ l | c }

    \Xhline{0.15em}
    \textbf{Measure} & \textbf{SCOPIL} \\
    \Xhline{0.15em}

    {Rwd} &  -21.84$\pm$2.76 \\ \cline{2-2}
    \Xhline{0.15em}
    
    {H } &  0.81$\pm$1.26 \\ \cline{2-2}
    \Xhline{0.15em}
    
    {F(H) } &  0.01$\pm$0.01 \\ \cline{2-2}
    \Xhline{0.15em}

    {V }  & 0.55$\pm$1.37 \\ \cline{2-2}
    \Xhline{0.15em}
    
    {F(V) } &  0.005$\pm$0.009 \\ \cline{2-2}
    \Xhline{0.15em}
    
    {C } & 0.35$\pm$0.12 \\ \cline{2-2}
    \Xhline{0.15em}
    
    {F(C) } & 0.011$\pm$0.003 \\ \cline{2-2}
    \Xhline{0.15em}

        {H+V+C } & 1.71$\pm$2.60 \\ \cline{2-2}
    \Xhline{0.15em}
    
    {F(H+V+C) } & 0.028$\pm$0.018 \\ \cline{2-2}
    \Xhline{0.15em}
    
    Length &  266.65$\pm$13.54 \\ \cline{2-2}
    \Xhline{0.15em}

    Steps &  31.41$\pm$3.66 \\ \cline{2-2}
    \Xhline{0.15em}
	
  \end{tabular}
  \label{table:multiconstraints_setting_results}
\end{center}
\end{table}

\textbf{Addressing mode collapse.} In this ($Simple$) setting, we observe an expert-level performance of SCOPIL. Specifically, it manages to surpass the reward obtained by the demonstrated trajectories, -16.16 and -18.30 respectively, while maintaining an extremely low number of violations. Notably, the average number of horizontal violations is 0.0, while the cyclic constraints violations are only 0.02 on average. This is also demonstrated in the training curves depicted in Figure \ref{fig:results_twomodes_scopil}, as well by the trajectories generated, depicted in Figure \ref{fig:two_modes_scopil_all_trajectories}, all  in Appendix \ref{sec:CurvesandTraj}. The significant point regarding SCOPIL  is that its performance increases when the trajectories are uniformly distributed across the different modes, in contrast to other methods which are prone to mode collapse.


\begin{table}
\begin{center}
\caption{Results for SCOPIL in \textit{Two-modes} setting.}
\scriptsize
  \begin{tabular}{ l | c }

    \Xhline{0.15em}
    \textbf{Measure} & \textbf{SCOPIL} \\
    \Xhline{0.15em}

    {Rwd} &  -16.16$\pm$0.25 \\ \cline{2-2}
    \Xhline{0.15em}
    
    {H } & 0.0$\pm$0.0 \\ \cline{2-2}
    \Xhline{0.15em}
    
    {F(H) } & 0.0$\pm$0.0 \\ \cline{2-2}
    \Xhline{0.15em}
    
    {C } & 0.02$\pm$0.0 \\ \cline{2-2}
    \Xhline{0.15em}
    
    {F(C) } & 0.003$\pm$0.00 \\ \cline{2-2}
    \Xhline{0.15em}

    {H+C } & 0.025$\pm$0.0 \\ \cline{2-2}
    \Xhline{0.15em}
    
    {F(H+C) } & 0.0035$\pm$0.0 \\ \cline{2-2}
    \Xhline{0.15em}
    
    Length &  264.43$\pm$131.22\\ \cline{2-2}
    \Xhline{0.15em}

    Steps &  24.95$\pm$11.94\\ \cline{2-2}
    
    \Xhline{0.15em}
  \end{tabular}
  \label{table:twomodes_setting_results}
\end{center}
\end{table}




\section{Related work}

 Safety and constraints-abiding behaviour is a long-term topic of interest in the RL realm.  It was formulated at first by \cite{Altman1999ConstrainedMD} through CMDPs.  Lagrangian relaxation plays an important role to most of the methods.
 CPO \cite{CPO} is a Constrained RL algorithm that provides guarantees for near-constraint satisfaction during training. CPO follows a trust region policy optimization approach complemented by Lagrangian relaxation, optimizing the advantage reward function while simultaneously aiming to minimize the advantage cost function. IPO \cite{Liu_Ding_Liu_2020} also employs Lagrangian relaxation integrated into PPO but it applies a logarithm barrier in the discounted cumulative cost function, imposing heavy penalization in the cases where any constraint is violated. Notably, the Lagrangian multiplier is replaced by a tunable (non-learnable) hyperparameter. On the other hand, \cite{pmlr-v119-stooke20a} introduces PID to address stability of updating the Lagrangian multiplier. Close to our work, \cite{pmlr-v155-ha21c} enhances SAC  by  dual gradient descent where the Lagrangian multiplier plays a crucial role in regulating the safety constraint function's magnitude throughout the training process. 
Another approach to incorporate both rewards and constraints in the optimization process was CVPO \cite{liu2022constrained}: It proposes a probabilistic inference approach, specifically an Expectation-Maximization method, along with Lagrangian Relaxation, which effectively tackles training instabilities. In a distinct direction, the authors of \cite{OptLayer} propose the integration of a safety layer, where actions are projected to the nearest feasible action w.r.t. the constraints.

Learning constraint-compliant policies exploiting expert demonstrations has been studied recently in the context of Inverse Safe and Constrained RL: ICRL\cite{ICRL}  aims to approximate the cost function and employs a binary classifier to differentiate between the demonstrated trajectories and those generated by the policy under learning.  Dual gradient descent and the PPO algorithm aim to optimize the RL objective and to minimize the cost by dynamically adjusting the  Lagrangian multiplier. Approximating the cost function and using it for optimization,  implies an increased training computational complexity. SPACE \cite{pmlr-v139-yang21i} aims to reduce constraint violations during execution and training: In iteration it 1) maximizes the reward advantage function using TRPO \cite{pmlr-v37-schulman15}, 2) projects the learnt policy onto a region around a baseline policy, and 3) aligns the learnt policy closely with the cost constraint region. SPACE needs a baseline policy, potentially utilizing expert demonstrations, and an explicit definition of the constraints. 

While these methods address safety and constraint satisfaction at varying degrees, they assume known cost functions and do not address the problem from an imitation learning perspective: In doing so, we show how the problem can be formulated in a probabilistic inference framework for constraints-abiding imitation learning, specifying a maximum entropy method for learning to adhere to constraints while optimizing the RL objective, by means of the Lagrangian relaxation  method. 

\section{Conclusions and further work}

This work  introduced an  constraints-abiding imitation learning method for learning  maximum entropy policies that comply with constraints demonstrated by means of expert trajectories executing a task, in a model-free way. The formulation of the problem and the objective function takes advantage of  results connecting performance to bounds for the KL-divergence between demonstrated and learned policies, and is connected to a probabilistic inference framework for reinforcement learning, incorporating the reinforcement learning objective and the  objective to abide by constraints in an entropy maximization setting. Based on this formulation, the proposed algorithm, SCOPIL, optimizes the learning objective with dual gradient descent, supporting effective and stable training. Experiments show that SCOPIL can learn effective policy models for constraints-abiding behaviour, in settings with multiple constraints of different types, accommodating different modalities of demonstrated behaviour, and with abilities to generalize.

Future work concerns extending SCOPIL in a safe method, dealing with hard constraints that the method must adhere to during training and  execution, while increasing the generalization abilities of the learnt policies in continuous state-action spaces.

\paragraph{Broader impact.} 
 The overarching objective is to develop and train agents capable of operating safely and addressing real-world problems effectively and in trustworthy manners. This is essential in many real-world safety critical applications (e.g. in traffic management, in robotic surgery, in providing summarized information of importance, in driving vehicles, etc) 
 where experts can demonstrate effective practices (of different modalities and behavioral patterns) without specifying explicitly constraints or cost functions. Adhering to expert or established practices in such settings is important, as it contributes to automation adoption. Furthermore, tuning effective policies to specific practices or making them transferable are important aspects, although not within the goals of his work. 
 
 While learning to abide by constraints is critical for applications, it is also important to understand properties (theoretical and practical) of imitation learning methods and connect these under unifying frameworks: In doing so, this work contributes to incorporating optimization with adherence to demonstrated constraints-abiding behaviour to probabilistic inference, formulating the constraints-abiding imitation learning problem.
 Regarding the proposed methods, e.g. SCOPIL, these must be enhanced with properties such as interpretability and provable safety. These are important implications that remain to be addressed by the community.





\bibliography{example_paper}
\bibliographystyle{icml2024}

\newpage
\appendix
\onecolumn

\section{Imitating constraints-abiding behaviour as probabilistic inference.}
\label{sec:ImitationAsProbInf}
This section aims to provide a connection of the problem formulation to reinforcements learning as probabilistic inference. This provides a very useful insight on the formulation towards learning safe and constrained policies via imitation learning, it provides a rigorous way to prove the correctness of the objective that we aim to optimize, and shows how incorporating rewards and constraints in the optimization process,  the proposed method unifies probabilistic inference approaches (e.g. CVPO \cite{liu2022constrained}), to maximum entropy imitation learning constraints-abiding methods.

First, following the notation and graphical model in \cite{RLasProbInf},  let $\mathcal{O}_t$ be a binary variable denoting whether the time step $t$ is optimal or not. Given the reward at time step $t$, $r(s_t, a_t)$, the distribution over this variable, is given by the following equation: $p(\mathcal{O}_t=1|s_t, a_t)=exp(r(s_t, a_t))$

In a similar way, let $\mathcal{C}_t$ be an additional binary variable denoting whether at time step $t$ constraints are satisfied. Given the cummulative cost at time step $t$, $C(s_t, a_t)$, the distribution over this variable, is given by the following equation: $p(\mathcal{C}_t=1|s_t, a_t)=1-exp(C(s_t, a_t))$.

Enhancing the graphical model with the $\mathcal{C}_t$ variables, we can have the posterior distribution over actions, when one acts optimally, $\mathcal{O}_t=1$, subsequently denoted $\mathcal{O}_t(s_t,a_t)$, and with respect to the constraints, $\mathcal{C}_t=1$, similarly denoted $\mathcal{C}_t(s_t,a_t)$, for all $t \in \{1,\dots,T\}$. In so doing we can compute the trajectory distribution $p(\tau, \mathcal{O}_{1:T}| \mathcal{C}_{1:T})$, when acting optimaly with respect to constraints, as follows:

\begin{equation}
p(\tau | \mathcal{O}_{1:T}, \mathcal{C}_{1:T}) \propto p(\tau, \mathcal{O}_{1:T}, \mathcal{C}_{1:T})= 
p(s_1)\prod_{t=1}^{T}p(\mathcal{O}_t(s_t,a_t)|\mathcal{C}_t(s_t,a_t))p(s_{t+1}|s_t,a_t)
\end{equation}

\assuming deterministic dynamics. According to Bayes' rule 

\begin{equation}
p(\mathcal{O}_t(s_t,a_t)|\mathcal{C}_t(s_t,a_t))=\frac{p(\mathcal{C}_t(s_t,a_t)|\mathcal{O}_t(s_t,a_t))p(\mathcal{O}_t(s_t,a_t))}{p(\mathcal{C}_t(s_t,a_t))}
\end{equation}

In an inverse reinforcement learning setting, given the occupancy measure $\rho_{\pi_E}$ of the demonstrated policy assumed optimal, it holds that 

\begin{equation}
p(\mathcal{C}_t(s_t,a_t)|\mathcal{O}_t(s_t,a_t))=\rho_{\pi_E}(s_t,a_t)[1-exp(C(s_t, a_t))]
\end{equation}

Therefore,
\begin{equation}
p(\mathcal{O}_t(s_t,a_t)|\mathcal{C}_t(s_t,a_t))=\rho_{\pi_E}(s_t,a_t)exp(r(s_t, a_t))
\end{equation}

Substituting $p(\mathcal{O}_t(s_t,a_t)|\mathcal{C}_t(s_t,a_t))$ from Eq. (17) in Eq. (14),

\begin{equation}
\begin{split}
& p(\tau, \mathcal{O}_{1:T},\mathcal{C}_{1:T})=\\
& p(s_1)\prod_{t=1}^{T}\rho_{\pi_E}(s_t,a_t)exp(r(s_t, a_t))p(s_{t+1}|s_t,a_t)=\\
&[p(s_1)\prod_{t=1}^{T}p(s_{t+1}|s_t,a_t)] \prod_{t=1}^{T} \rho_{\pi_E}(s_t,a_t)\prod_{t=1}^{T}exp(r(s_t, a_t))
\end{split}
\end{equation}

The first term in Eq. (18) is constant for trajectories that are dynamically feasible, given deterministic dynamics, while the second and the third terms show that the probability of a trajectory increases proportionally to the occupancy measure of the demonstrated policy and exponentially to the reward. 

Therefore, we aim at optimal trajectories that are with respect to the dynamics and demonstrated policy:

\begin{equation}
\begin{split}
p(\tau| \mathcal{O}_{1:T}, \mathcal{C}_{1:T}) 
\propto \mathds{1}[p(\tau) \neq 0]\prod_{t=1}^{T}\rho_{\pi_E}(s_t,a_t)\prod_{t=1}^{T}exp(r(s_t, a_t))
\end{split}
\end{equation}

The indicator function indicates that the trajectory is feasible given the dynamics. 

We aim  for an approximation $\pi(a_t|s_t)$ that adheres perfectly to the policy generating the demonstrated trajectories, such that the trajectory distribution

\begin{equation}
\begin{split}
\hat p(\tau) \propto \mathds{1}[p(\tau) \neq 0]\prod^T_{t=1}\pi(a_t|s_t)
\end{split}
\end{equation}

matches the distribution given in Eq. (19). Therefore, we need to have
\begin{equation}
\begin{split}
D_{KL}(\hat p(\tau) || p(\tau)) = -\mathbb{E}_{\tau \sim \hat p(\tau)}[log p(\tau)- log \hat p(\tau)]= 0
\end{split}
\end{equation}
\\

This, applying the $log$ to $p(\tau)$ and $\hat p(\tau)$, results to the objective of maximizing

\begin{equation}
\begin{split}
&\mathbb{E}_{\tau \sim \hat p(\tau)}[\sum_{t=1}^T[log(\rho_{\pi_E}(s_t,a_t))+r(s_t, a_t)]-log \pi(a_t|s_t)]=\\
&\sum_{t=1}^{T} \mathbb{E}_{(s_t,a_t) \sim \hat p(s_t,a_t)}([log(\rho_{\pi_E}(s_t,a_t))+r(s_t, a_t)]-log \pi(a_t|s_t))=\\
&\sum_{t=1}^{T} \mathbb{E}_{(s_t,a_t) \sim \hat p(s_t,a_t)}([log(\rho_{\pi_E}(s_t,a_t))]-\beta log \pi(a_t|s_t)) + \\
&\sum_{t=1}^{T} \mathbb{E}_{(s_t,a_t) \sim \hat p(s_t,a_t)}([r(s_t, a_t)]-(1-\beta)log \pi(a_t|s_t))= \\
&\sum_{t=1}^{T} \mathbb{E}_{(s_t,a_t) \sim \hat p(s_t,a_t)}[log(\rho_{\pi_E}(s_t,a_t))]+ \beta \mathbb{E}_{s_t \sim \hat p (s_t)}[\mathcal{H}(\pi(a_t|s_t)]+ \\
&\sum_{t=1}^{T} \mathbb{E}_{(s_t,a_t) \sim \hat p(s_t,a_t)}[r(s_t, a_t)]+(1-\beta)\mathbb{E}_{s_t \sim \hat p (s_t)}[\mathcal{H}(\pi(a_t|s_t)]=\\
&-(\sum_{t=1}^{T} \mathbb{E}_{(s_t,a_t) \sim \hat p(s_t,a_t)}[-log(\rho_{\pi_E}(s_t,a_t))]- \beta \mathbb{E}_{s_t \sim \hat p (s_t)}[\mathcal{H}(\pi(a_t|s_t)])\\
&+ \sum_{t=1}^{T} \mathbb{E}_{(s_t,a_t) \sim \hat p(s_t,a_t)}[r(s_t, a_t)]+(1-\beta)\mathbb{E}_{s_t \sim \hat p (s_t)}[\mathcal{H}(\pi(a_t|s_t)]
\end{split}
\end{equation}

The entropy term in  Eq. (22) has been split into the two terms by a factor $\beta \in [0,1]$: The second term is the  maximum entropy reinforcement learning objective and the first incorporates the KL Divergence between the occupancy measures of the two policies, which has to be minimized. To make the role of the first term in Eq. (22) clearer, the first term is approximated by  the $D_{KL}(\pi_E||\pi)$. By also reversing the order of terms, the final objective becomes

\begin{equation}
\begin{split}
\sum_{t=1}^{T} \mathbb{E}_{(s_t,a_t) \sim \hat p(s_t,a_t)}[r(s_t, a_t)]+(1-\beta)\mathbb{E}_{s_t \sim \hat p (s_t)}[\mathcal{H}(\pi(a_t|s_t)] - (D_{KL}(\pi_E||\pi)- \beta \mathbb{E}_{s_t \sim \hat p (s_t)}[\mathcal{H}(\pi(a_t|s_t)]) 
\end{split}
\end{equation}

This allows us (a) to respect modalities in the demonstrated trajectories, and (b) have a clear role for parameter $\beta \in [0,1]$, which balances the role of policy entropy between the two sub-objectives: Both terms aim to maximize the entropy, the first with respect to the optimization objective, while the second with respect to the demonstrated constraints-abiding trajectories.

\section{Learning to satisfy constraints while maximizing the entropy: A side note}
\label{sec:SideNote}
It is worth pointing out a key difference between Eq. (\ref{eq:entr_coef_obj}) and (\ref{eq:g_algo}) which both are objectives for updating the corresponding Lagrangian multipliers $\alpha$ and $\lambda$, respectively. 

As it can be observed  the target entropy $\Bar{\mathcal{H}}$ in Eq. (\ref{eq:entr_coef_obj}) has a positive sign while the policy entropy has a negative sign. In contrast, $\delta$ in Eq. (\ref{eq:g_algo}), which can be considered to be the target there, has a negative sign and  the metric distance has a positive sign. This is because the Lagrangian relaxation formulation of SAC with regard to $\alpha$ aims to maximize the policy entropy, where only the values  $\geq \Bar{\mathcal{H}}$ are considered feasible, and the target is the optimal one. On the other hand, according to Eq. (\ref{eq:dual_func}), in our formulation, we aim to minimize $D_{KL}$ while maintaining the policy entropy considered by SAC, where only the values $\leq \delta$ are considered feasible, and $\delta$ is the optimal one. 

For reference (and comparison with Eq. (\ref{eq:dual_func})), we write the Lagrangian relaxation of SAC as presented in \cite{haarnoja2019soft}:
\begin{equation}
\begin{split} \label{eq:sac_lagr_relax}
    & \max_{\pi \in \Pi} \mathbb{E}_{\tau \sim \rho_{\pi}} \left[ \sum_{t=0}^{T} r(s_t,a_t) \right] \\
    & \mathrm{s.t.} \ \mathbb{E}_{(s_t,a_t) \sim \rho_{\pi}} \left[ -\log \left( \pi \left( a_t | s_t \right) \right) \right] \geq  \Bar{\mathcal{H}}, \ \ \forall t
\end{split}
\end{equation}
Based on the above, we can conclude that SCOPIL, built on SAC, uses   but \textit{Double} Dual Gradient Descent, where 
the Lagrangian is optimized with respect to: (a) the expected discounted reward (maximization), (b) the policy's $\pi_{\theta}$ entropy (maximization), (c) the distance measure between policies $\pi_{\theta}$ and $\pi_E$ for constraint adherence (minimization), and (d) the Lagrangian multipliers $\lambda$ and $\alpha$ (maximization).

\newpage
\section{Experimental results: Training curves and ball trajectories}

\subsection{Training curves and representative trajectories in the $Simple$ setting.}

\label{sec:CurvesandTraj}
\begin{figure}[ht]
  \centering
  \includegraphics[width=0.55\linewidth]{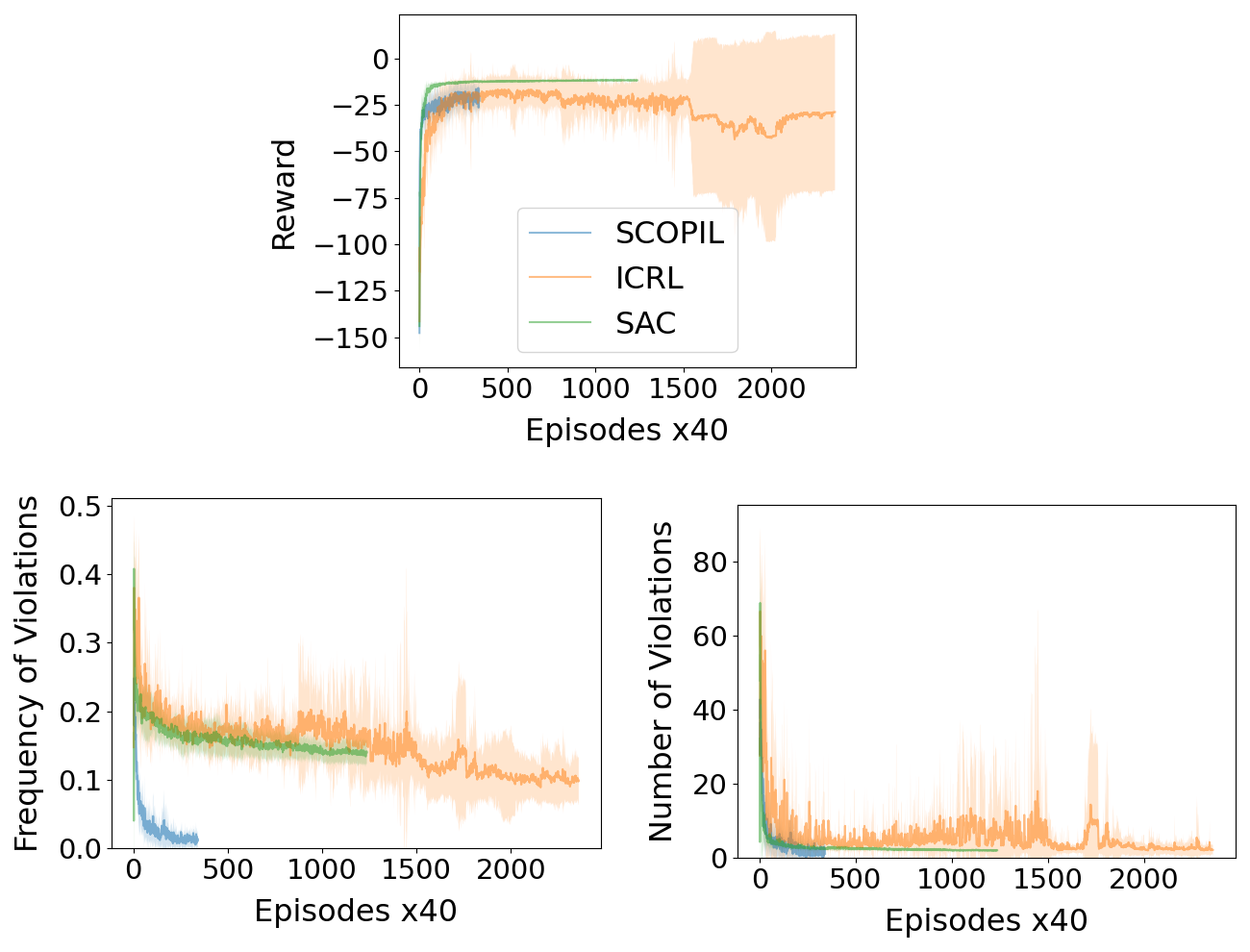}
  \vspace*{-4mm}
  \caption{Training curves for SCOPIL, ICRL, and SAC in \textit{Simple} setting. (1st row): Reward, (2nd row): Frequency and number of horizontal line constraint violations, (3rd row): Frequency and number of cyclic constraint violations.}
  \label{fig:main_results}
  \Description{}
\end{figure}

\begin{figure}[ht]
  \centering
  \includegraphics[width=0.6\linewidth]{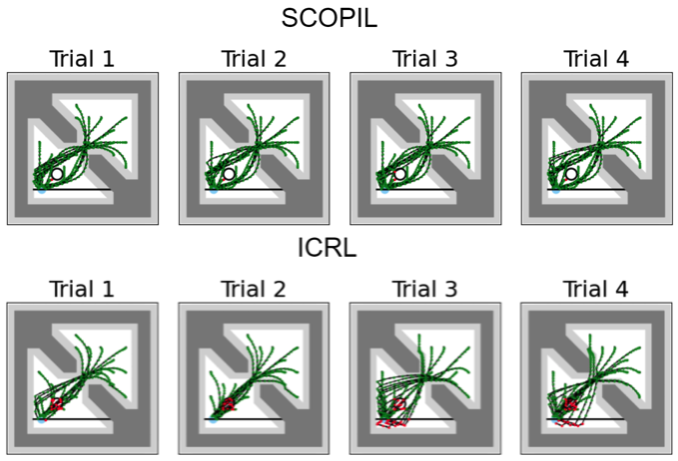}
  \vspace*{-4mm}
  \caption{Ball trajectories for the first 4 training trials (independent experiments) of SCOPIL and ICRL in \textit{Simple} setting.}
  \label{fig:results_trajectories_scopil_icrl}
  \Description{}
\end{figure}

\newpage
\subsection{Training curves and representative trajectories in the $Multiple-constraints$ setting.}

\begin{figure}[ht]
  \centering
  \includegraphics[width=0.5\linewidth]{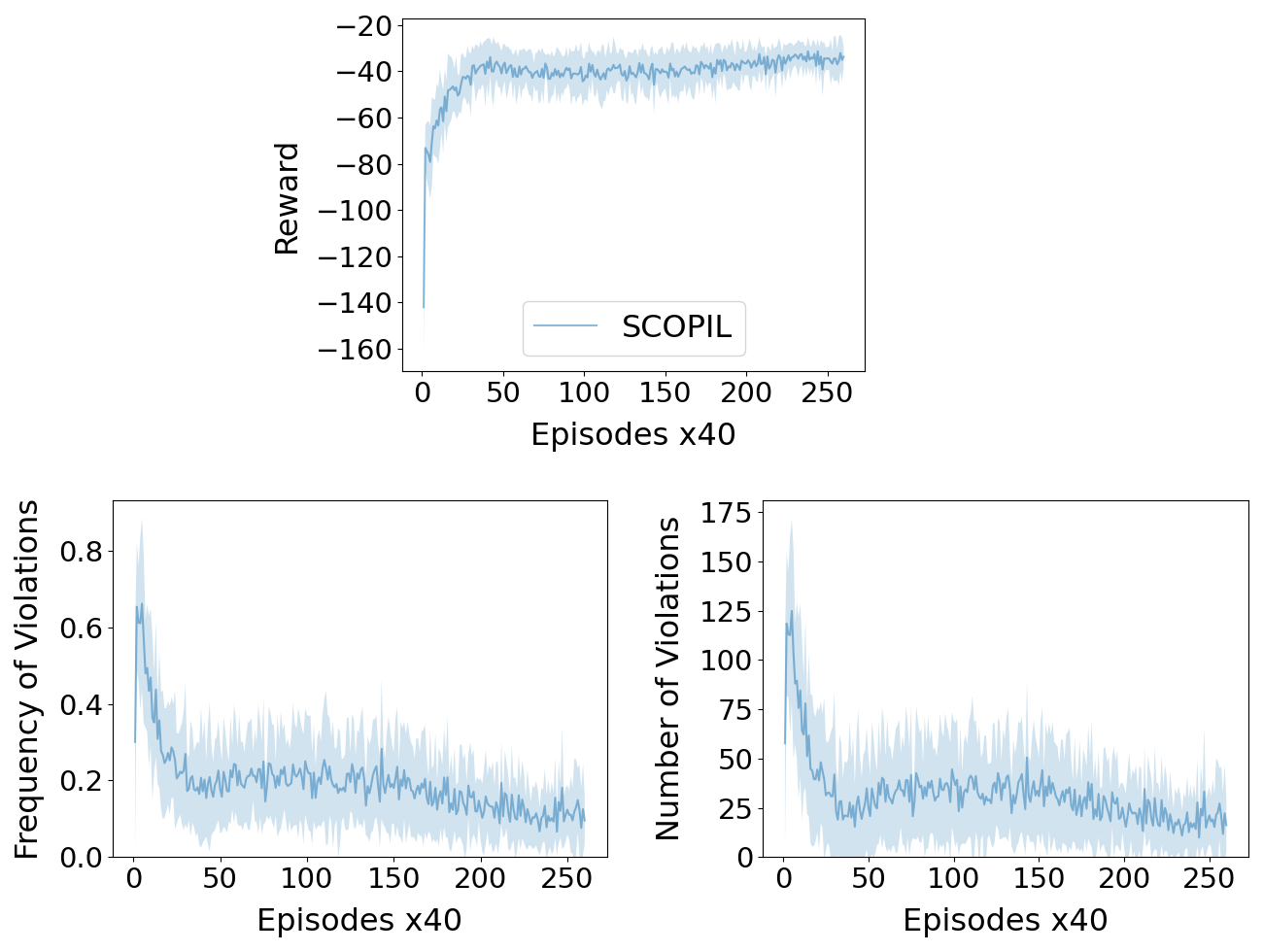}
  \vspace*{-4mm}
  \caption{Training curves for SCOPIL in \textit{Multiple-constraints} setting. (1st row): Reward, (2nd row): Frequency and number of horizontal line constraint violations, (3rd row): Frequency and number of vertical line constraint violations, (4th row): Frequency and number of cyclic constraints violations.}
  \label{fig:scopil_results_multiconstraints}
  \Description{}
\end{figure}

\begin{figure}[ht]
  \centering
  \includegraphics[width=0.6\linewidth]{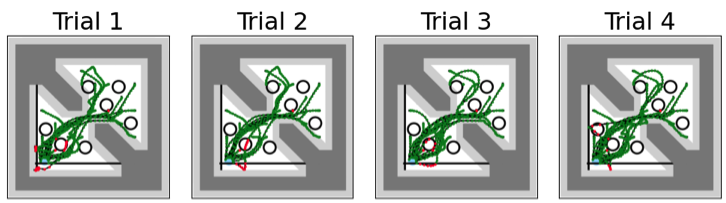}
  \vspace*{-4mm}
  \caption{Ball trajectories for the first 4 training trials (independent experiments) of SCOPIL in \textit{Multiple-constraints} setting.}
  \label{fig:multiconstraints_scopil_all_trajectories_scopil_only_4}
  \Description{}
\end{figure}

\newpage
\subsection{Training curves and representative trajectories in the $Two-modes$ setting.}

\begin{figure}[ht]
  \centering
  \includegraphics[width=0.6\linewidth]{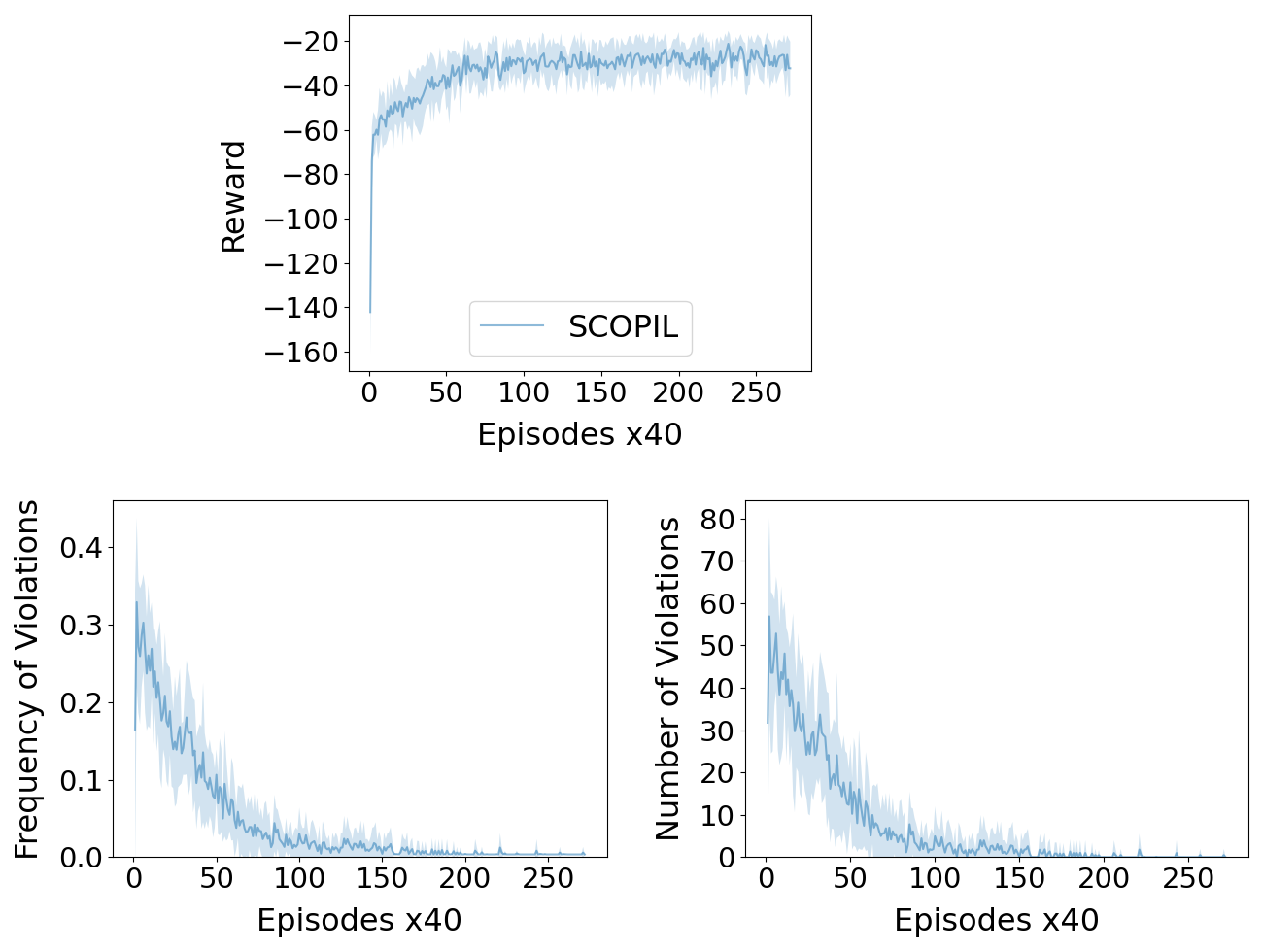}
  \vspace*{-4mm}
  \caption{Training curves for SCOPIL in \textit{Two-modes} setting. (1st row): Reward, (2nd row): Frequency and number of horizontal line constraint violations, (3rd row): Frequency and number of cyclic constraints violations.}
  \label{fig:results_twomodes_scopil}
  \Description{}
\end{figure}

\begin{figure}[h]
  \centering
  \includegraphics[width=0.45\linewidth]{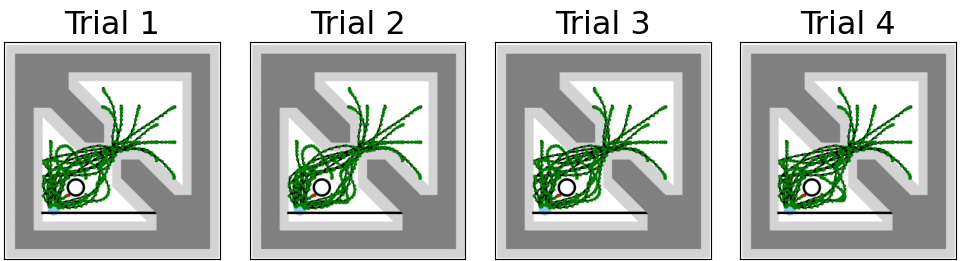}
  \caption{for the first 4 training trials (independent experiments) of SCOPIL in \textit{Two-modes} setting.}
  \label{fig:two_modes_scopil_all_trajectories}
  \Description{}
\end{figure}

\newpage
\section{Ablation Study}
\label{sec:Ablation}
This ablation study aims to study the effects of (a) tuning the Lagrangian multiplier as part of the dual gradient descent (DGD) process, (b) splitting the entropy term between the RL objective and the constraints adherence term, and (c) satisfying the RL objective with another algorithm.

To investigate these aspects, this section reports on SCOPIL with a non-effective dual gradient descent (denoted S-DGD), without the entropy in the constraints term (denoted by S-EiC), and with a non-effective dual gradient descent without the entropy in the constraints term (denoted S-DGD-EiC). S-DGD maximizes the objective of SAC with respect to constraints but with a constant Lagrangian multiplier $\lambda$. The same happens for S-DGD-EiC but without the entropy in the constraints term. For the third aspect, we report results from the method resulting by replacing SAC with PPO (denoted PPO+DGD). 
Results report on the average reward, average number of violations for each of the constraints type and for all types, the average frequency of violating constraints per trajectory step, as well as the length and the steps per trajectory, all for the $Simple$ setting.

\begin{table}[h]
\scriptsize
\begin{center}
\caption{Results for S-DGD, S-EiC, S-DGD-EiC, and PPO+DGD in \textit{Simple} setting. \textit{S} and \textit{P} stand for \textit{SCOPIL} and \textit{PPO}, respectively.}

  \begin{tabular}{ l | c |c | c | c | c}

    \Xhline{0.15em}
    \textbf{Measure} & \textbf{SCOPIL} &  \textbf{S-DGD} & \textbf{S-EiC} & \textbf{S-DGD-EiC} & \textbf{PPO+DGD}\\
    \Xhline{0.15em}

    {Rwd} & -13.46$\pm$0.29 & -13.22$\pm$0.19 & -13.76$\pm$0.21 & -13.41$\pm$0.36 & -10.94$\pm$0.11 \\ \cline{2-6}
    \Xhline{0.15em}
    
    {H } & 0.045$\pm$0.95 &  0.12$\pm$0.16 & 0.005$\pm$0.016 & 0.10$\pm$0.18 & 0.08$\pm$0.07 \\ \cline{2-6}
    \Xhline{0.15em}
    
   {F(H) } & 0.002$\pm$0.005 & 0.006$\pm$0.008 & 0.000$\pm$0.001 & 0.004$\pm$0.006 & 0.005$\pm$0.006 \\ \cline{2-6}
   \Xhline{0.15em}
    
    {C } & 0.042$\pm$0.04 & 0.02$\pm$0.00 & 0.03$\pm$0.01 & 0.025$\pm$0.000 & 0.64$\pm$0.17 \\ \cline{2-6}
    \Xhline{0.15em}
    
    {F(C) } & 0.004$\pm$0.001 & 0.003$\pm$0.000 & 0.004$\pm$0.001 & 0.003$\pm$0.000 & 0.05$\pm$0.01 \\ \cline{2-6}
    \Xhline{0.15em}

    {H+C } & 0.087$\pm$0.126 & 0.1475$\pm$0.163 & 0.035$\pm$0.021 & 0.125$\pm$0.184 & 0.72$\pm$0.138 \\ \cline{2-6}
    \Xhline{0.15em}
    
    {F(H+C) } & 0.006$\pm$0.006 & 0.01$\pm$0.008 & 0.004$\pm$0.001 & 0.007$\pm$0.006 & 0.05$\pm$0.01 \\ \cline{2-6}
    \Xhline{0.15em}
    
    Length & 255.32$\pm$1.57 & 256.58$\pm$3.09 & 255.50$\pm$2.42 & 256.13$\pm$3.01 & 245.06$\pm$2.13 \\
    \Xhline{0.15em}

    Steps & 19.83$\pm$0.41 & 19.48$\pm$0.27 & 20.26$\pm$0.31 & 19.76$\pm$0.52 & 16.25$\pm$0.16 \\
    
  \end{tabular}
  \label{table:simple_setting_results_ablation_study}
\end{center}
\end{table}

Results reported in Table \ref{table:simple_setting_results_ablation_study} show that all variations, except S-EiC, are not as  effective  in satisfying the constraints as SCOPIL. S-EiC is very competitive in finding a constraints-adhering policy regarding violations of all types of constraints, as well as with regard to the frequency of violations. All other methods, except PPO+DGD report very good results for the cyclic constraints.

\begin{figure}[h]
  \centering
  \includegraphics[width=0.55\linewidth]{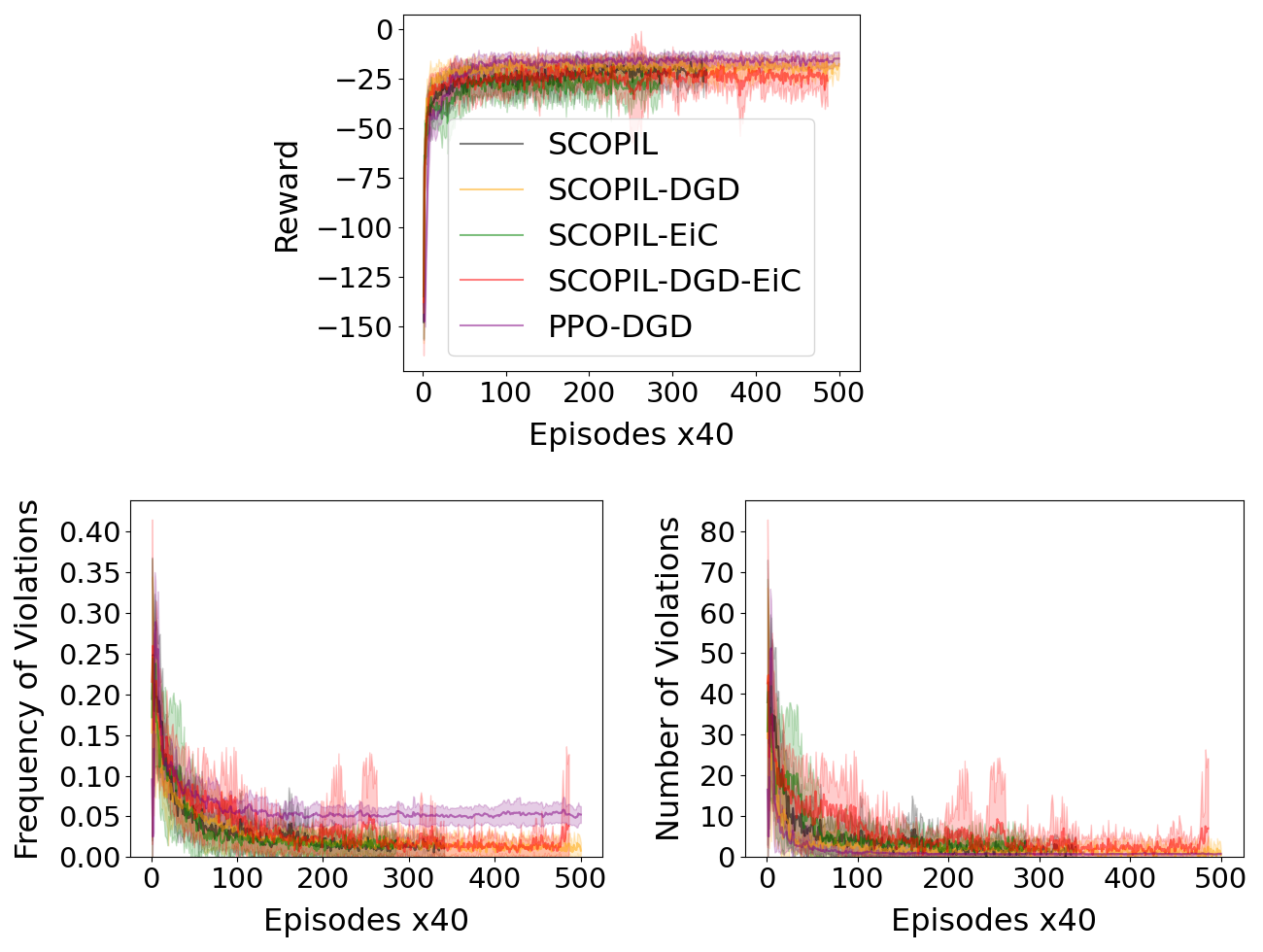}
  \vspace*{-4mm}
  \caption{Training curves for SCOPIL, SCOPIL-DGD, SCOPIL-EiC, SCOPIL-DGD-EiC, and PPO+DGD in \textit{Simple} setting. \\ \textit{-DGD} and \textit{-EiC} stand for \textit{without Dual Gradient Descent} and \textit{without Entropy in Constraints term}, respectively.  (1st row): Reward, (2nd row): Frequency and number of horizontal line constraint violations, (3rd row): Frequency and number of cyclic constraint violations.}
  \label{fig:ablation_results}
  \Description{}
\end{figure}

 Training curves in Figure \ref{fig:ablation_results} show that  methods, except PPO+DGD and SCOPIL-DGD-EiC, compete well during training, reducing the constraint violations and their frequency in fast pace.

\begin{figure}[h]
  \centering
  
\includegraphics[width=0.4\linewidth]{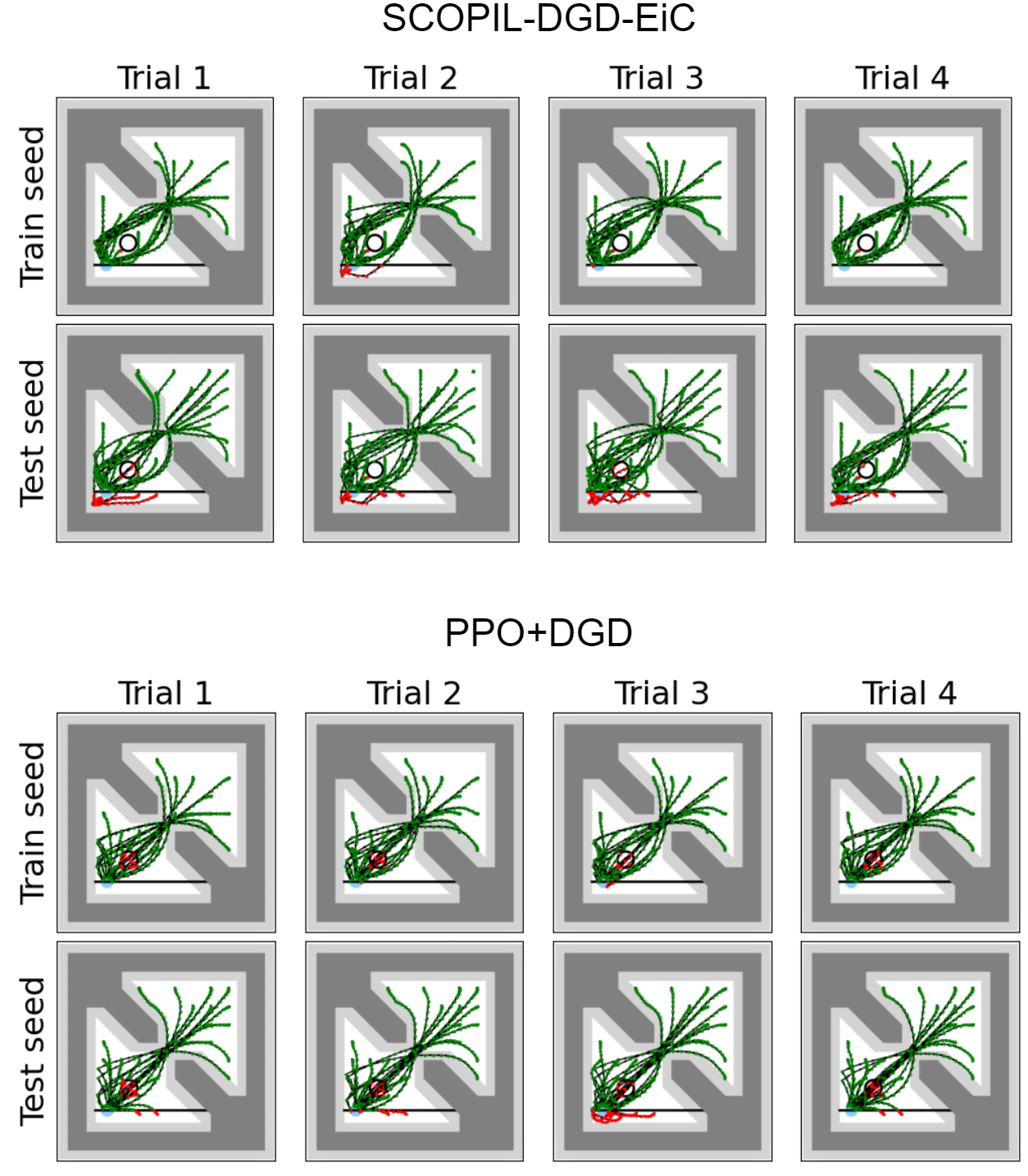}
  \vspace*{-4mm}
  \caption{Ball trajectories for the first 4 training trials (independent experiments) of SCOPIL-DGD-EiC and PPO+DGD in \textit{Simple} setting.}
  \label{fig:ablation_study_results_trajectories_scopil-dgd-eic_ppo+dgd}
  \Description{}
\end{figure}

All the above phenomena and results, are due to the reasons explained  in Section \ref{sec:results} with regard to SAC. This is further evidenced by results demonstrated in Figure \ref{fig:ablation_study_results_trajectories_scopil-dgd-eic_ppo+dgd} with regard to PPO+DGD, where the absence of SAC results into mode collapse. 

\begin{figure}[h]
  \centering
  \includegraphics[width=0.4\linewidth]{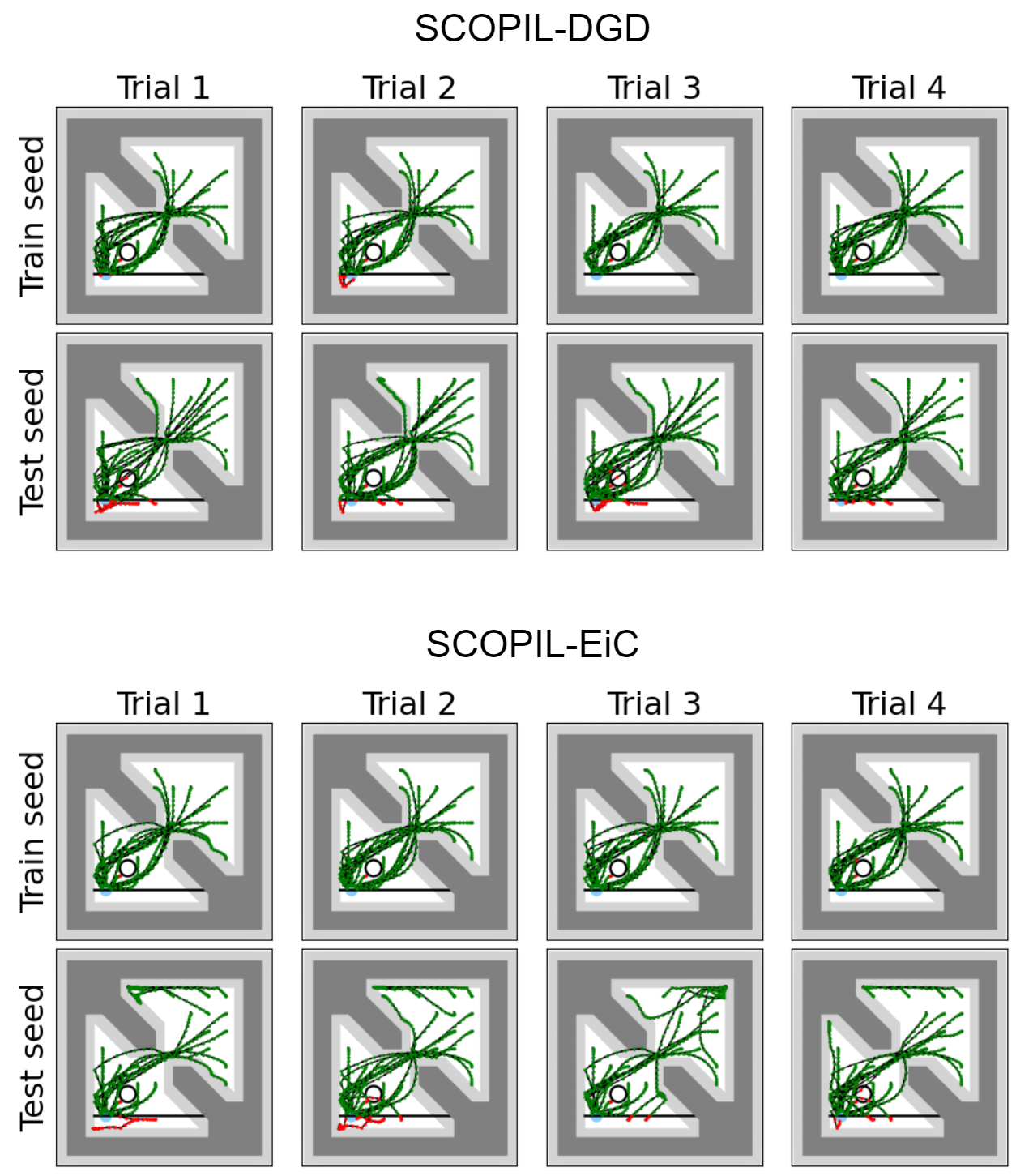}
  \vspace*{-4mm}
  \caption{Ball trajectories for the first 4 training trials (independent experiments) of SCOPIL-DGD and SCOPIL-EiC in \textit{Simple} setting.}
  \label{fig:ablation_study_results_trajectories_scopil-dgd_scopil-eic}
  \Description{}
\end{figure}

Methods without the entropy in the constraints term of the objective (i.e. S-EiC and  S-DGD-EiC) manage to adhere effectively to constraints since, they ``force" close match to the occupancy measure of the demonstrated policy, reducing the ability of the method to generalize beyond samples seen. Delving more on the details of these results, Figures \ref{fig:ablation_study_results_trajectories_scopil-dgd_scopil-eic} and \ref{fig:ablation_study_results_trajectories_scopil-dgd-eic_ppo+dgd} show trajectories generated by the ablated versions of SCOPIL both, with initial states from the distribution of initial states in the demonstrations (Train seed) and with novel initial states (Test seed). 
S-EiC ``train seed" trajectories in Figure \ref{fig:ablation_study_results_trajectories_scopil-dgd_scopil-eic}  and S-DGD-EiC ``train seed" trajectories in Figure \ref{fig:ablation_study_results_trajectories_scopil-dgd-eic_ppo+dgd} show adherence to constraints.

\begin{figure}[h]
  \centering
  
\includegraphics[width=0.4\linewidth]{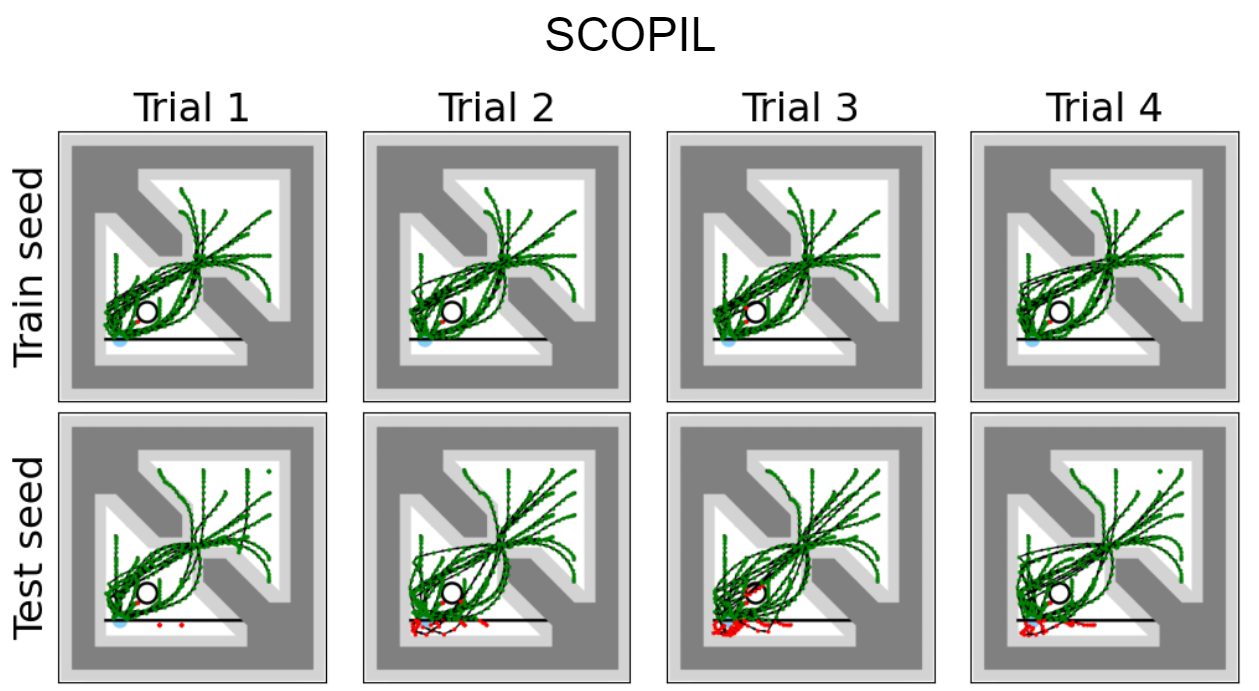}
  \vspace*{-4mm}
  \caption{Ball trajectories for the first 4 training trials (independent experiments) of SCOPIL in \textit{Simple} setting.}
  \label{fig:scopil-trajectories-simple}
  \Description{}
\end{figure}

 However, ``test seed" trajectories for S-EiC show that they are not effective to reaching the goal state, moving the goal towards the right-upper part of the board (without violating constraints), while ``test seed" trajectories for S-DGD-EiC demonstrate a mode collapse issue. This shows the ineffectiveness of these ablated methods to infer effective policies from the demonstrations. Figure \ref{fig:scopil-trajectories-simple} provides the SCOPIL generated trajectories given train and test seeds: While the violation of constraints increases with the test seeds, the policy generates consistent trajectories that are quite effective, given the demonstrations provided and the number of SCOPIL training episodes.

\newpage
\section{Training specifications.}

In this section, we provide the details of training for SCOPIL and ICRL in Table (\ref{table:scopil_hyperparameters}) and (\ref{table:icrl_hyperparameters}), respectively. For SAC, we use the same hyperparameters with SCOPIL, except from $\lambda_{0}$. For PPO+DGD we use the same hyperparameters with ICRL, that is, the first 7 reported in Table (\ref{table:icrl_hyperparameters}). However, we use the same $\lambda_{0}$ and $\lambda$ learning rate with SCOPIL.

The source code and the datasets are available at: \url{https://github.com/AILabDsUnipi/SCOPIL}.

\begin{table}[h]
\centering
\caption{SCOPIL hyperparameters.}
\begin{tabular}{c | c}
\Xhline{0.15em}
Batch size                 & 256       \\ \hline
$\eta$                     & 0.0003    \\ \hline
$\kappa$                   & 0.002     \\ \hline
$\gamma$                   & 0.99      \\ \hline
$\epsilon$                 & 0.005     \\ \hline
Buffer memory size         & 1\,000\,000 \\ \hline
\# steps                   & 1\,000\,000 \\ \hline
$\lambda_{0}$              & 1.05      \\ \hline

\multicolumn{2}{c}{\textbf{Neural-network architecture}}\\ \hline
Layers                     & 2         \\ 
Neurons per layer          & [32, 32]  \\ 
\Xhline{0.15em}
\end{tabular}
\label{table:scopil_hyperparameters}
\end{table}

\begin{table}[ht]
\centering
\caption{ICRL hyperparameters.}
\begin{tabular}{ c | c }
\Xhline{0.15em}

Batch size                       & 64 \\ \hline

\multicolumn{2}{c}{\textbf{Policy, $\pi_{\phi}$}}\\ \hline
Learning rate                    & 0.0003 \\ 
PPO target KL                    & 0.02 \\
Reward-GAE $\gamma$              & 0.99 \\
Reward-GAE $\lambda$             & 0.95 \\
Entropy bonus $1/\beta$          & 0.0 \\
Network architecture             & [64,64] \\ \hline

\multicolumn{2}{c}{\textbf{Lagrangian, $\lambda$}}\\ \hline
Initial value                    & 1.0  \\
Learning rate                    & 0.1  \\
Budget                           & 0.0  \\ \hline

\multicolumn{2}{c}{\textbf{Constraint function, $\zeta_{\theta}$}}\\ \hline
Network architecture             & [40,40] \\
Learning rate                    & 0.01  \\
Backward iterations              & 20    \\
Regularizer weight               & 0.5   \\
Max forward KL $\epsilon_{F}$    & 10    \\
Max backward KL $\epsilon_{B}$   & 2.5   \\ \hline

Expert rollouts                  & 100   \\ \hline
Rollout length                   & 400   \\ \hline
\# steps per iteration           & 200000 \\ \hline
\# iterations                    & 40    \\
\Xhline{0.15em}
\end{tabular}
\label{table:icrl_hyperparameters}
\end{table}




\end{document}